\documentclass[sigconf]{acmart}
\AtBeginDocument{%
  \providecommand\BibTeX{{%
    \normalfont B\kern-0.5em{\scshape i\kern-0.25em b}\kern-0.8em\TeX}}}

\copyrightyear{2025} 
\acmYear{2025} 
\setcopyright{cc}
\setcctype{by}
\acmConference[KDD '25]{Proceedings of the 31st ACM SIGKDD Conference on Knowledge Discovery and Data Mining V.2}{August 3--7, 2025}{Toronto, ON, Canada}
\acmBooktitle{Proceedings of the 31st ACM SIGKDD Conference on Knowledge Discovery and Data Mining V.2 (KDD '25), August 3--7, 2025, Toronto, ON, Canada}
\acmDOI{10.1145/3711896.3736843}
\acmISBN{979-8-4007-1454-2/2025/08}

\usepackage{microtype}
\usepackage{graphicx}

\usepackage{booktabs} % for professional tables

\usepackage{float} 
\usepackage{subcaption}

\usepackage{caption}
\usepackage{mathtools}
\usepackage{amsfonts}
\usepackage{dsfont}

\usepackage{url}

\usepackage{xcolor}
\usepackage{url}
\usepackage{multirow} 
\usepackage{amsmath,amsthm}
\usepackage{booktabs}
\usepackage{graphicx}
\usepackage{wrapfig}
\usepackage{algorithmic}
\usepackage{algorithm}

\newcommand{\ie}{\textit{i.e.}}
\newcommand{\eg}{\textit{e.g.}}

\acmConference[KDD '25]{the 31th ACM SIGKDD Conference on Knowledge Discovery and Data Mining}{August 3--7, 2025}{Toronto, Canada}
\usepackage{multirow}
\usepackage{amsthm}
\usepackage{amsmath}

\begin{document}

\title{AnomalyGFM: Graph Foundation Model for Zero/Few-shot Anomaly Detection}
% with Normal Information }

\author{Hezhe Qiao$^{\dagger}$}
\affiliation{%
  \institution{School of Computing and Information
Systems\\ Singapore Management University}
    \country{Singapore}
}
\email{hezheqiao.2022@phdcs.smu.edu.sg}

 \author{Chaoxi Niu$^{\dagger}$}
\affiliation{%
 \institution{Australian Artificial Intelligence Institute\\ University of Technology Sydney}
 \country{Sydney, Australia }
}
 \email{chaoxi.niu@student.uts.edu.au}

 \author{Ling Chen}
\affiliation{%
 \institution{Australian Artificial Intelligence Institute\\ University of Technology Sydney}
 \country{Sydney, Australia }
}
 \email{ling.chen@uts.edu.au}

 \author{Guansong Pang}
\authornote{Corresponding author: Guansong Pang,  ${\dagger}$
indicates equal contribution.}
\affiliation{%
 \institution{School of Computing and Information
Systems\\ Singapore Management University}
   \country{Singapore}
}
 \email{gspang@smu.edu.sg}

% \renewcommand{\shortauthors}{Anonymous Author, et al.}

\begin{abstract}

Graph anomaly detection (GAD) aims to identify abnormal nodes that differ from the majority of the nodes in a graph, which has been attracting significant attention in recent years.  Existing generalist graph models have achieved remarkable success in different graph tasks but struggle to generalize to the GAD task. This limitation arises from their difficulty in learning generalized knowledge for capturing the inherently infrequent, irregular and heterogeneous 
% class imbalance, the non-uniform distribution of
abnormality patterns
% , and the challenge of learning fine-grained invariant features across
in graphs from different domains.
% , which hinder them from serving as foundational models for GAD. 
To address this challenge, we propose \textbf{AnomalyGFM}, a GAD-oriented graph foundation model that supports zero-shot inference and few-shot prompt tuning for GAD in diverse graph datasets. One key insight is that graph-agnostic representations for normal and abnormal classes are required to support effective zero/few-shot GAD across different graphs.
% by leveraging the residual feature of normal class and abnormal class.
% We first empirically reveal the residual features of a node to its neighbor is uniformly maintained to ensure consistency across the graphs and discrimination in normality and abnormality, contrasting to the original feature which facilitates the learning of unified prompts. Specifically, 
Motivated by this, AnomalyGFM is pre-trained to align data-independent, learnable normal and abnormal class prototypes with node representation residuals (\ie, representation deviation of a node from its neighbors). 
% we construct two trainable prompts drawn from normal distribution corresponding to the normal class and abnormal class to learn the normal and abnormal
The residual features essentially project the node information into a unified feature space where we can effectively measure the abnormality of nodes from different graphs in a consistent way. This provides a driving force for the learning of graph-agnostic, discriminative prototypes for the normal and abnormal classes, which can be used to enable zero-shot GAD on new graphs, including very large-scale graphs. If there are few-shot labeled normal nodes available in the new graphs, AnomalyGFM can further support prompt tuning to leverage these nodes for better adaptation.
% respectively on the training datasets. In the inference step, we use similarity measurement with two prompts to achieve three kinds of inference including (1) Zero-shot inference, (2) Few-shot inference, and (3) Subgraph inference.
Comprehensive experiments on 11 widely-used GAD datasets  with real anomalies, covering social networks, finance networks, and co-review networks, demonstrate that AnomalyGFM significantly outperforms state-of-the-art competing methods under both zero- and few-shot GAD settings.
% can effectively leverage the residual features's discrimination across datasets to construct a graph foundation model for GAD. (2) the effectiveness and generalization of AnomalyGFM on zero-shot and few-shot settings. 
Code is available at \renewcommand\UrlFont{\color{blue}} \url{https://github.com/mala-lab/AnomalyGFM}.

% \gs{pls use prototypes rather than prompts for the two learned representations, since they are learned in the pre-training stage. the representations learned during the tuning stages are more suitable for the term `prompt'}

\end{abstract}


% \ccsdesc[300]{Anomaly detection}
% \ccsdesc{Anomaly detection}

% \keywords{Do, Not, Us, This, Code, Put, the, Correct, Terms, for,


\maketitle

\section{Introduction}

Graph anomaly detection (GAD) aims to identify anomalous nodes within a graph that deviate significantly from others. It has found broad applications in areas such as financial fraud detection, transaction analysis, and social networks \cite{akoglu2015graph, ma2021comprehensive,pang2021deep,qiao2024deep}. Most existing GAD methods, either supervised or unsupervised methods, follow a paradigm of training and inference on the same graph, making the trained models struggle to generalize to new/unseen graphs due to the distribution shift between the training and testing graphs \cite{qiao2024truncated,liu2021anomaly,ding2019deep}. Further, this type of methods can become inapplicable in applications where graph data is not accessible during training due to data privacy or other data access issues. 
Generalist/foundation models for graphs have achieved remarkable progress in tackling these challenges in general tasks such as node classification, graph classification, and link prediction \cite{sun2023all, li2024zerog, zhao2024all, zi2024prog, fang2024universal}, but they are difficult to generalize to the GAD task. The main reason is that the generalized knowledge learned in these generalist models through generic pre-training or tuning methods inclines to the prevalent regular patterns, which differ significantly from the inherently infrequent, irregular, and heterogeneous abnormality patterns across the graphs
% which significantly hinders the advancement of GAD techniques and poses challenges to developing a robust foundation model
\cite{qiao2024deep, liu2024arc, lin2024unigad}.

 \begin{figure}
 \centering
 \includegraphics[width=2.8in,height=2.1in]{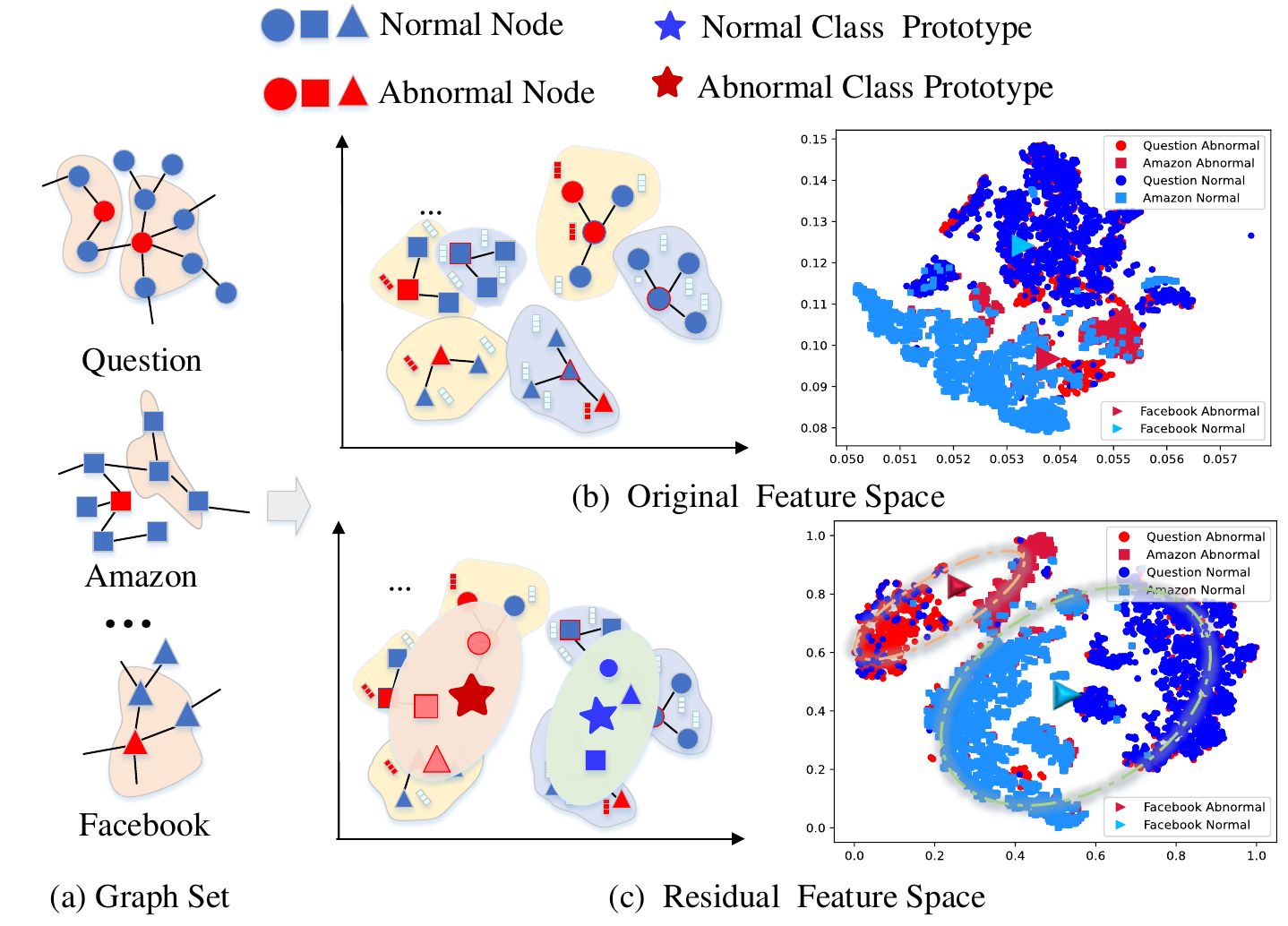}\\
 \caption{Given a set of graph datasets consisting of normal nodes (blue) and abnormal nodes (red) as in \textbf{(a)}, AnomalyGFM aims to learn graph-agnostic prototypes for normal and abnormal nodes in a projected residual feature space. Compared to the original feature space where normal and abnormal patterns are irregularly distributed and differ significantly in different datasets as in \textbf{(b)}, the residual feature space learned by AnomalyGFM in \textbf{(c)} is universally effective in separating anomalies from normal nodes in different graphs. In (b) and (c), AnomalyGFM is trained on Facebook \cite{xu2022contrastive} and evaluated directly on Amazon \cite{dou2020enhancing} and Question \cite{platonov2023critical}; both feature spaces are learned by AnomalyGFM. 
 % The t-SNE visualizations of node representations for the original features and residual features of the real dataset on Amazon, Question, and Facebook (training dataset), are shown in Fig. (b) and Fig. (c), respectively. The original features, derived from different datasets, exhibit poor classification performance, see Fig. (b). In contrast, the residual features are uniformly structured, ensuring clear discrimination across datasets from different domains, and the normal and abnormal representations of Facebook are located in the center of the corresponding cluster. This highlights the effectiveness of studying the unified prototype of normal and abnormal nodes compared to the original representations, see Fig. (c).
 % \gs{changes are required to b and c. they are hard to read. you may use the same color for each class, and different shapes for nodes from different datasets.}
 % \gs{do we need to emphasize ``Normal/abnormal Residual Feature'' in the top legend? this information is in the caption of subfigure (c). Suggest to unify the first two columns in the top legend}
 % \gs{how about further replace 'Graph A' with `Facebook', `Graph B' with `Amazon', and `Graph C' with `Question'? pls add citation for the three datasets in the caption.}
}
 \label{fig:example}
 \vspace{-1em} 
\end{figure}

Very recently, there have been some efforts, such as ARC \cite{liu2024arc} and UNPrompt \cite{niu2024zero}, dedicating to generalist models for GAD.
% to address this limitation, 
Despite the effectiveness on some datasets, these methods fail to provide a 
% unified 
% abnormality measurement across the graphs and the solution
universal GAD model
that versatilely supports zero-shot inference and sample-efficient tuning in different application scenarios. For instance, UNPrompt \cite{niu2024zero} is designed exclusively for zero-shot inference, while ARC \cite{liu2024arc} is limited to few-shot tuning-based GAD within the same domain.
Therefore, a foundational model being capable of effectively capturing abnormalities across graphs from different domains and supporting both zero-shot and few-shot GAD is still lacking.

To fill this gap, in this paper, we introduce \textbf{AnomalyGFM}, a GAD-oriented graph foundation model (GFM) that is capable of addressing both zero-shot and few-shot scenarios effectively for GAD. One key insight is that to avoid overfitting the pre-training graph data, graph-agnostic representations for normal and abnormal classes are required to support effective zero/few-shot GAD across different graphs. Motivated by this, AnomalyGFM is pre-trained to align data-independent, learnable normal and abnormal class prototypes with \textit{node representation residuals} (\ie, representation deviation of a node from its neighboring nodes). Compared to the original feature space where normal and abnormal patterns are irregularly and heterogeneously distributed in different datasets (see Fig. \ref{fig:example} (b)),
the residual features essentially project the node information into a unified feature space where we can effectively measure the abnormality of nodes from different graphs in a consistently identical way, as shown in Fig. \ref{fig:example} (c). This is because normal nodes, regardless of which graphs they are from, are expected to have small residual/deviation from its neighboring nodes, whereas such residuals for abnormal nodes are large \cite{qiao2024truncated,liu2024arc}.
% The residual feature can essentially make the abnormality be measured in a consistent way by projecting the original feature into a unified feature space, 
% As illustrated in Fig. \ref{fig:example}, the residual features are universally effective in separating anomalies from normal nodes across graphs in the representation space, reducing domain dependence, see in Fig. \ref{fig:example} (b), while the original features where normal and abnormal patterns are irregularly distributed across the different graphs, see in Fig. \ref{fig:example} (c). This separating anomalies from normal nodes in the residual feature across the graphs makes it easier to identify the prototype of the normal and abnormal class
% and facilitates the establishment of a foundation model that can measure the abnormality using the prototypes. Therefore, the goal of
By learning to align such residual features with the learnable normal and abnormal class prototypes,
AnomalyGFM distills the residual feature-based anomaly discriminability into the two graph-agnostic prototypes, facilitating strong generalization to GAD in new graphs without any further tuning/training. Furthermore, if there are few-shot labeled normal nodes available in the new graphs, AnomalyGFM can further support graph prompt tuning to leverage these nodes for better adaptation.
% that align with the center of normal and abnormal residual feature respectively, 
% helping the model trained on a source graph can be effectively generalized to the new graphs.

To summarize, the contributions of AnomalyGFM are three-fold:
\begin{itemize}
\item  We propose a new paradigm for GAD, aiming to pre-train a GAD-oriented GFM that achieves strong zero-shot generalization and supports sample-efficient graph prompt tuning for the GAD task.
% and demonstrate that the residual features can serve as the graph-agnostic discriminative feature which makes the abnormality in the graph can be measured in a consistent way.

\item  We argue that learning graph-agnostic representations for normal and abnormal classes is a key ingredient of a GAD-oriented GFM. To justify this, 
% effectively leverage the graph-agnostic representations-residual feature, 
we propose AnomalyGFM, the very first GAD-oriented GFM with strong zero-shot and few-shot generalization abilities. It is pre-trained to learn discriminative, graph-agnostic class prototypes with normal and abnormal residual features, and it supports few-shot graph prompt tuning for better adaptation.

% \item  AnomalyGFM can be used to enable zero-shot GAD on new graphs, including very large-scale graphs. If there are few-shot labeled normal nodes available for the new graphs, AnomalyGFM
% can further support prompt tuning to leverage these nodes for better adaptation.

\item A comprehensive benchmark on both zero-shot and few-shot settings using 11 real-world GAD datasets is established, on which i) AnomalyGFM performs significantly better than the state-of-the-art unsupervised, supervised, and generalist GAD methods, and ii) AnomalyGFM can scale up to very large graphs.

\end{itemize}

\section{Related Work}
\subsection{Graph Anomaly Detection} 
Existing GAD methods can be generally classified into unsupervised and supervised methods \cite{ma2021comprehensive, qiao2024deep}, depending on the assumption regarding the presence of normal and anomaly labels. The fully supervised approaches have recently achieved remarkable progress by leveraging both labeled normal and abnormal nodes \cite{tang2022rethinking, liu2021pick, gao2023addressing, wang2023open}. These methods are primarily designed to improve message aggregation in graph neural networks (GNNs) by reducing heterophilic edges from both spectral and spatial perspectives, effectively mitigating the over-smoothing problem in GAD \cite{tang2023gadbench, qiao2024deep}. Fully supervised methods are effective in GAD as they treat the task as an imbalanced classification problem. However, the requirement for both normal and abnormal labels significantly hinders their applications to scenarios where labeled nodes are difficult to obtain.

Unsupervised methods, which typically assume the absence of both normal and abnormal labels, have garnered significant attention due to their more practical setting assumption on data labels. They generally incorporate some conventional techniques such as reconstruction \cite{ding2019deep}, one-class classification \cite{wang2021one, zhou2021subtractive,qiao2024truncated}, contrastive learning \cite{liu2021anomaly, pan2023prem}, and adversarial learning \cite{ding2021inductive} into graph learning to capture the normal patterns within the graph and then assign an anomaly score for each node based on its deviation from the normal patterns. However, these methods still follow the paradigm of training and inference on the same graph, making them struggle to generalize to new/unseen graphs due to the distribution shift between training and testing set.

% Unsupervised methods can operate on data without labels, but their constraints are often too strict for real-world applications. 
%  Due to the strict constraints of the unsupervised methods and the fact that normal labels are generally easier to obtain than abnormal labels in real-world applications \cite{qiao2024generative}, semi-supervised methods that consider partially known normal labels have been gradually introduced. These methods are more aligned with practical scenarios. Some unsupervised methods can be adapted to semi-supervised settings by directly applying normal pattern extraction techniques to the clean normal samples~\cite{ding2019deep, wang2021one, liu2021anomaly}. However, these methods still follow the paradigm of training and inference on the same graph making them struggle to generalize to the new graph from different domains.

% GGAD \cite{qiao2024generative} is the first semi-supervised method to address the scenario where only a subset of normal labels is available. It generates pseudo-anomalous nodes from partially labeled normal nodes to provide effective negative samples for training a discriminative one-class classifier. 

\subsection{Foundation Models for GAD}
Generalist models have recently achieved significant progress on non-graph data by leveraging large pre-trained models to facilitate generalized pattern recognition in diverse downstream tasks, such as generalist anomaly detection on images \cite{zhou2023anomalyclip,zhu2024toward}. However, applying these methods to graph data remains highly challenging due to the absence of such pre-trained models \cite{sun2023all, liu2023graphprompt, wen2023voucher} on graph data.
% Despite significant advancements in foundation models across various domains, the development of GFMs is still in its early stages. Recent studies have shown initial successes in specialized areas like knowledge graphs \cite{} and molecular graphs 
The main challenge in designing a GFM is capturing invariance across diverse graphs while mapping them into a shared representation space to enable positive transfer between training and inference \cite{maoposition,beaini2023towards,galkin2023towards}. Currently, most GFM models use prompt learning to enable the knowledge transfer across graphs for general graph tasks \cite{liu2023graphprompt, fang2024universal}.
% For example, GraphPrompt \cite{liu2023graphprompt} builds a pre-training and prompting framework designed for both graph-level and node-level classification tasks. It learns a prompt that helps downstream tasks identify and leverage the most relevant knowledge from the pre-trained model in a task-specific way.
% While GFMs have made significant progress in general graph tasks, 
However,
they still struggle to generalize to the GAD task due to the inherently infrequent, irregular, and heterogeneous
abnormality patterns in GAD datasets \cite{liu2024arc, niu2024zero, qiao2024deep}. Therefore, 
% some generalist methods that versatilely support zero-shot inference and sample-efficient tuning have been proposed for graph anomaly detection very recently by training on source graphs and evaluating on test graphs
some recent efforts attempt to devise the GFMs for GAD \cite{niu2024zero, liu2024arc}.
ARC \cite{liu2024arc} is one of such methods, a fine-tuning GFMs for GAD, enabling a ``one-for-all” GAD model through a fine-tuning mechanism. 
% It introduces an ego-neighbor residual graph encoder that learns node embeddings sensitive to abnormalities. During inference, a cross-attentive in-context anomaly scoring module is used to predict node abnormality by leveraging a few-shot set of normal samples.
UNPrompt \cite{niu2024zero} is the first zero-shot generalist GAD method that learns generalized neighborhood prompts, allowing latent node attribute predictability to serve as a generalized anomaly measure.
% across diverse datasets without requiring fine-tuning on the target dataset.
Different from ARC and UNPrompt, which address the GAD-oriented GFMs partially under few-shot or zero-shot settings, AnomalyGFM can effectively capture abnormalities across graphs from different domains while supporting both zero-shot inference and few-shot prompt fine-tuning/inference. This offers more versatile abilities for a GFM-based GAD approach.

\section{The Proposed AnomalyGFM}
\subsection{Preliminaries}
\noindent \textbf{Notation and Conventional GAD.}
An attributed graph is denoted by $\mathcal{G}=(\mathcal{V}, \mathcal{E})$ where $\mathcal{V}=\{v_1,v_2,...,v_N\}$ represents the node set and $\mathcal{E}$ represents the edge set. The graph can also be denoted as $\mathcal{G}=(\mathbf{A}, \mathbf{X})$ where $\mathbf{X}=\left[\mathbf{x}_1, \mathbf{x}_2, \ldots, \mathbf{x}_N\right]^T \in \mathbb{R}^{N \times \tilde{d}}$ is the attribute of the nodes and $\mathbf{A} \in \mathbb{R}^{N \times N}$ is the adjacent matrix of the graph with $\mathbf{A}_{ij} = 0$  representing there is no edge between node $v_i$ and node $v_j$ and $A_{ij} = 1$ otherwise. In GAD, we denote the normal node set as ${\mathcal V}_n$ and abnormal node set ${\mathcal V}_a$, corresponding to the anomaly labels ${\bf{y}} \in {\{ 0,1\} ^N}$ with ($y_i=0$ denotes a normal node $v_i$ and $y_i=1$ denotes abnormal). 
Typically,  the number of normal nodes is significantly larger than the abnormal nodes in the GAD datasets, \ie, $\left| {{{\mathcal V}_n}} \right| \gg \left| {{{\mathcal V}_a}} \right|$. The conventional GAD is formulated as learning a scoring function $f: \mathcal{G} \rightarrow \mathbb{R}$ and making inference on the same $\mathcal{G}$, in which $f$ is expected to assign every node an anomaly score such that $f(v) < f(v^{\prime})$, $\forall v\in \mathcal{V}_{n}, v^{\prime} \in \mathcal{V}_{a}$.

\noindent \textbf{Zero-shot GAD.} 
Different from the conventional GAD that trains and inferences on the same graph, zero-shot GAD aims to train a generalist model on $\mathcal{G}_{Train}$ and then apply the learned model to the new test graphs $\mathcal{G}_{Test}$ without any further tuning/retraining.

\noindent \textbf{Few-shot GAD.}
In the few-shot GAD setting, the model is fine-tuned on $\mathcal{G}_{Test}$ with small labeled nodes and then applied to the rest of the unlabelled nodes in the test graph $\mathcal{G}_{Test}$. We denote the labeled node in the test graph as $\mathcal{V}_l^{Test}$ while the unlabeled nodes in the test graph are denoted as $\mathcal{V}_{u}^{Test}=\mathcal{V}^{Test}\setminus\mathcal{V}_l^{Test}$. 

% Here we only consider the case of having few-shot labeled normal nodes since the normal node labels are easier/less costly to obtain than the abnormal ones \cite{liu2024arc}.

\noindent \textbf{Feature Unification.}
Due to the feature dimension difference across the graphs, we need to align the node features/attributes into a shared feature space to ease the feature heterogeneity among the graphs. Following the previous studies \cite{liu2024arc, niu2024zero}, we employ singular value decomposition (SVD), a simple, efficient dimensionality reduction technique that can approximately preserve the distance relationship of the data by projecting high-dimensional data into a low-dimensional subspace for unification, For each dataset $\mathbf{X}^{(i)}$ with dimension $d^{(i)}$, 
it can be formulated as:
\begin{equation}
{\mathbf{X}^{(i)} \in \mathbb{R}^{N^{(i)} \times d^{(i)}} \xrightarrow[\text { Projection }]{\text { Feature }} \tilde{\mathbf{X}}^{(i)} \in \mathbb{R}^{N^{(i)} \times d^{\prime}}},
\end{equation}
where $d^\prime$ is the common dimensionalities and $N^{(i)}$ is the number of node for graph $\mathcal{G}_i$

\noindent \textbf{GNN for Representation Learning.}
Due to its simplicity and effectiveness, GNN is employed to learn the representation of each graph which can be formulated as follows,
\begin{equation}
{
\mathbf{H}^{(\ell)}=\operatorname{GNN}\left(\mathbf{A}, \mathbf{H}^{(\ell-1)} ; \mathbf{W}^{(\ell)}\right),
}
\end{equation}
where ${\bf{H}}^{(\ell)} \in \mathbb{R}^{N\times h^{(l)}}$, ${\bf{H}}^{(\ell-1 )}\in\mathbb{R}^{N\times h^{(l-1)}}$ are the representations of all $N$ nodes in the ${(\ell)}$-th layer and ${(\ell -1)}$-th layer, respectively, $h^{(l)}$ is the dimension of ${(\ell)}$-th layer, ${\bf{W}}^{(\ell)}$ are the learnable parameters, and ${\bf{H}}^{(0)} = \tilde{\mathbf{X}}$. ${{{\bf{H}}} = \left\{ {{{\bf{h}}_1},{{\bf{h}}_2}, \ldots ,{{\bf{h}}_N}} \right\}}$ is a set of representations of $N$ nodes in the last GNN layer.
% , with ${{{\bf{h} \in{\mathbb{R}}}^d} }$.
In this paper, we adopt a 2-layer GCN to model the graph  due to its efficient and simple architecture.

\subsection{Approach Overview}
\begin{figure*}
\centering
% Requires \usepackage{graphi
\includegraphics[width=6.45in,height=2.1in]{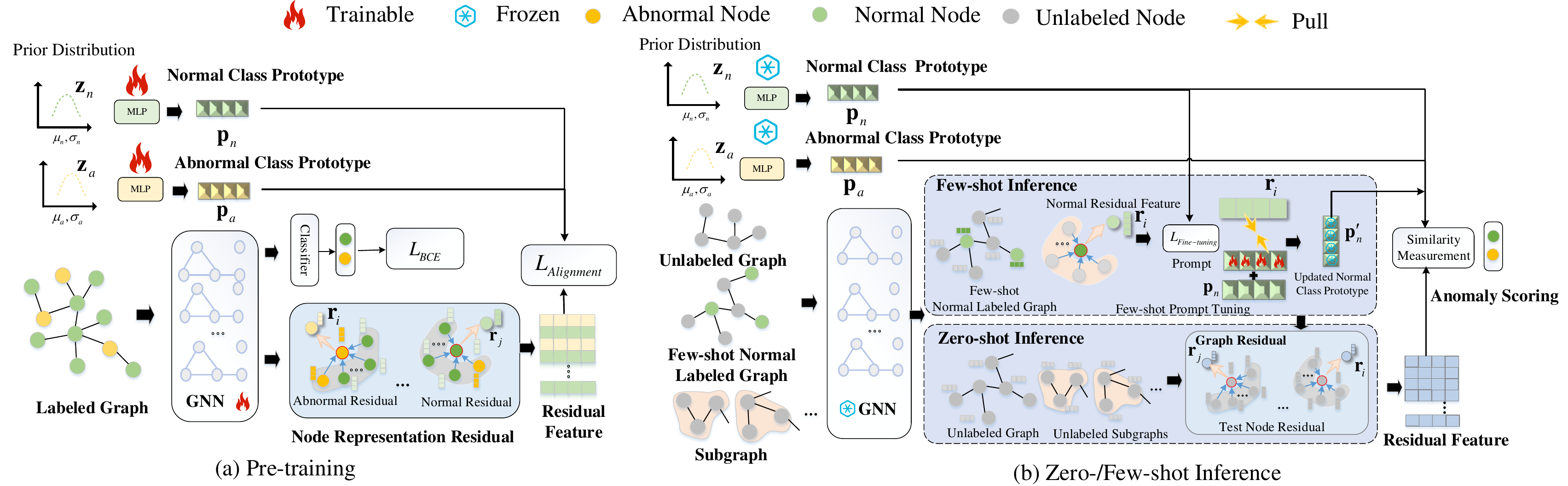}\\
\caption{Overview of AnomalyGFM. (a) During the pre-training, AnomalyGFM aims to align two learnable class-level prototypes with the representation residuals of normal and abnormal nodes to learn discriminative graph-agnostic prototypes on an auxiliary labeled graph. 
% the model is trained on a labeled graph where the GNN is employed to learn node representations using the BCE loss function, and the normal and abnormal residual features are used to guide the learning of two data-independent prototypes through the alignment loss. 
(b) During inference,  the anomaly score is determined in a consistent way using similarity between node representation residual and prototypes where all learnable parameters are frozen. Given a target graph with few-shot labeled normal nodes, these nodes are used to better adapt the pre-trained normal class prototypes to the target graph through a one-class prompt tuning method.
% \gs{change caption in (b) to Zero-/Few-shot Inference pls}
}
\label{fig:framework}
\vspace{-0.5em} 
\end{figure*}

% \gs{pls give a concise introduction to the key intuition of AnomalyGFM before telling the procedures.}
AnomalyGFM aims to distill discriminative node residual features in a unified feature space into two graph-agnostic prototypes for the normal and abnormal classes, facilitating the establishment of a foundation model that supports effective zero/few-shot GAD across graphs.
Specifically, AnomalyGFM consists of pre-training based on the prototype alignment, few-shot graph prompt fine-tuning, and zero-shot/few-shot inference, as shown in Fig. \ref{fig:framework}. In the pre-training step, we aim to align the data-independent and learnable normal class and abnormal class prototypes with the nodes' residual features to achieve the distillation of graph-agnostic discriminative class-level representations. In the few-shot graph prompt fine-tuning, the labeled normal nodes' residual features are utilized to guide a better adaptation of the normal class prototype. In the inference, we devise zero-shot inference on new graphs, including subgraph-based inference for very large-scale graphs, and few-shot inference, both of which compute anomaly scores by measuring the similarity between the residual features of nodes and the two class prototypes.

\subsection{Pre-training via Prototype Alignment in a Unified Feature Space}

\noindent \textbf{Node Representation Residual.} 
Node representation residual can be seen as the deviation at the representation level between the target node and its neighboring nodes.
The deviation in the representation between the target node and its neighbors is expected to be small for normal nodes, regardless of the graph domain, while the deviation for abnormal nodes is relatively large. Therefore, the node representation residual offers a consistent method to effectively capture heterogeneous abnormalities across graphs.
Assume the node representation outputted by the GNNs as $\mathbf{h}_i \in \mathbb{R}^d$, where $d$ is the dimension of representation, the residual feature $\mathbf{r}_i \in \mathbb{R}^{d}$ for the node $v_i$ is formally defined as follows:
\begin{equation}
\label{eq:residual}
    {
\mathbf{r}_i=\mathbf{h}_i-\frac{1}{\left|\mathcal{N}\left(v_i\right)\right|} \sum_{v_j \in \mathcal{N}\left(v_i\right)} \mathbf{h}_j,
    }
\end{equation}
where $\mathcal{N}\left(v_i\right)$ represents the neighbor set of node $v_i$. As a graph-agnostic representation, the node representation residual will be assigned to each node as its new feature in the unified feature space.

\noindent \textbf{Graph-agnostic Prototype Alignment.} 
To exploit the anomaly discriminability in node representation residuals, we learn two data-independent learnable prototypes for the two classes that
% which both draw from a prior distribution, \eg, Gaussian distribution, to
align with the normal and abnormal residual features, respectively. The alignment essentially distills the residual feature-based anomaly discriminability into the two graph-agnostic prototypes, facilitating a strong generalization to GAD in new graphs.
Formally, we integrate two learnable mapping functions ${\bf p} _n =\Phi({\bf{z}}_n; \Theta_n)$, where ${\bf{z}}_n,{\bf p} _n \in \mathbb{R}^T$ and ${\bf p} _a = \Phi({\bf{z}}_a; \Theta_a)$, where ${\bf{z}}_a,  {\bf p} _a \in \mathbb{R}^T$ to map their initializations (\eg, prototypes initialized with vectors drawn from a prior distribution such as Gaussian distribution) into the prototype space to serve as normal class prototype $\mathbf{p}_{n}$ and abnormal class prototype $\mathbf{p}_{a}$, respectively. Both $\Theta_n$ and $\Theta_a$ are the learnable parameters, and $T$ is the dimension of the learnable prototype, which is set to match the dimensionality of the node representation $d$. The Gaussian prior distributions ${\bf{z}}_n$ and ${\bf{z}}_a$ are uniformly defined as $\mathcal{N}(\mu, \sigma)$ for the normal and abnormal classes, where ${\mu}$ is the mean of the distribution and $\sigma$ is its standard deviation.

% \noindent \textbf{Pre-training.} 
The pre-training of AnomalyGFM is performed by aligning the two learnable prototypes with the abnormal residual features and normal residual features respectively using an alignment loss function $L_{Alignment}$ defined as follows:
\begin{equation}
\label{eq:alignment}
    {
{L_{Alignment{\rm{ }}}} = \sum\limits_{i = 1}^{\left| {{{\mathcal V}}} \right|} {{I_{{y_{i = 0}}}}} \left\| {{{\bf{r}}_i} - {{\bf{p}}_n}} \right\|_2^2 + {I_{{y_{i = 1}}}}\left\| {{{\bf{r}}_i} - {{\bf{p}}_a}} \right\|_2^2
    }\, ,
\end{equation}
% where ${\bf{r}}_n = \frac{1}{{\left| {{{{\mathcal V}}_n}} \right|}}\sum\limits_{{v_i} \in {{\mathcal V}_n}} {{{\bf{r}}_i}} $  and ${\bf{r}}_a = \frac{1}{{\left| {{{\mathcal V}_a}} \right|}}\sum\limits_{{v_i} \in {{\mathcal V}_a}} {{{\bf{r}}_i}} $ are the average residual feature of normal nodes and abnormal nodes respectively.
where ${I_{{y_{i = 0}}}}$ is an indicator function, which takes the value of 1 when the condition is satisfied.
This distills the graph-agnostic discriminability in the residual feature space into the two prototypes. To enhance the anomaly discriminability in the node representation residual, we utilize a standard binary classification loss $L_{BCE}$ to learn discriminative representations for the normal and abnormal nodes in the original node representation space:
% in learning more effective and distinguishable representations. 
\begin{equation}
{L_{BCE}} = -\sum\limits_{ i=1}^{\left| {{{\mathcal V}}} \right|} {{y_i}} \log \left( {{p_i}} \right) + \left( {1 - {y_i}} \right)\log \left( {1 - {p_i}} \right),
\end{equation}
where ${{p_i} = {f_\theta }\left( {{\bf h}_i} \right)}$ and $f_\theta(\cdot)$ is a Multi-Layer Perceptron (MLP) classifier with training parameters $\theta$, which maps the representation to the probability for a node being an anomaly.  Finally, we jointly optimize the ${L_{Alignment}}$ and $L_{BCE}$ via the following total loss :
% $L_{total}$ is defined as,
\begin{equation}
{
{L_{total{\rm{ }}}} = {L_{BCE}} + \alpha {L_{Alignment{\rm{ }}}}
},
\end{equation}
where $\alpha$ is a hyperparameter to control the importance of the proposed alignment module. Note that the pre-training is performed on auxiliary graph datasets rather than target graph datasets.

\subsection{Few-shot Prompt Tuning}
% \gs{pls move the relevant prompt tuning content from Sec. 4.2 to here}
To enable the few-shot prompt tuning in the scenarios where few-shot labeled normal nodes are available in the test/target graph, we utilize these limited nodes to have a prompt tuning of AnomalyGFM in the prototype learning module while keeping the GNN parameters frozen. 
% Note that only normal nodes are considered in the few-shot setting, as normal labels are typically easier to obtain than abnormal ones in the real application \cite{liu2024arc, qiao2024generative}.

\noindent \textbf{Normal Prototype-based Prompting.} To fully leverage the labeled normal nodes during the fine-tuning step, we introduce a small learnable prompt into the normal class prototype to better align it with the normal node representation residuals on the test graph during a prompt fine-tuning step. In addition, we also add an adaptation layer that offers more learning flexibility to the tuning.  They are designed to refine the pre-trained normal class prototype ${\bf{p}}_n$ to a new normal prototype $\bf{p}_n^\prime$, ensuring a better alignment with the normal class residual feature in the target graphs.  Formally, the new normal prototype ${\bf{p}^\prime_n}$ is defined as:
% \begin{equation}
% {\bf{p}^\prime_n} = g\left( {{{\bf{p}}_n};\phi ,\Psi } \right) = {{\bf{p}}_n} + {f_\phi }({{\bf{p}}_n}) + \Psi,
% \end{equation}

\begin{equation}
{\bf{p}^\prime_n}  = {{\bf{p}}_n} + {g}({{\bf{p}}_n;\phi}) + \Psi_{\mathbf{p}_n},
\end{equation}
where $\Psi_{\mathbf{p}_n} \in \mathbb{R}^d$ is the learnable prompt while ${g}({{\bf{p}}_n;\phi})$ is the added adaptation layer with parameters $\phi$, both of which are of the same dimensionality as ${{\bf{p}}_n} $. The learned prompts and the adapted prototype complements the pre-trained normal prototype in the target graph, so we fuse these three vectors to obtain the updated normal class prototype ${\bf{p}^\prime_n}$.
% , which
% that adapts the pre-trained normal class, serving as complementary information to the original normal class prototype for better adaptation to the target dataset.

\noindent \textbf{One-class Prompt Tuning.} During the prompt tuning, the original learnable parameters of AnomalyGFM 
% of the original GNN $\bf{W}$ and the learnable mapping function for the prototypes $\Theta_{n}$ and $\Theta_a$ 
are all frozen. The few-shot labeled normal nodes are solely utilized to learn the new parameters $\phi$ and $\Psi$ for refining the normal class prototype. Specifically, given some labeled normal node $\mathcal{V}_l^{Test}$ in the test graph, we first calculate the residual feature ${\bf{r}}_i$ of the labeled normal node $v_i \in \mathcal{V}_l^{Test}$ using Eq. \ref{eq:residual}. 
Then the graph prompt is optimized using the following one-class tuning loss $L_{{{pt}}}$, aiming to minimize the Euclidean distance between the residual features of labeled normal node ${\bf r}_i$ 
and the new trainable normal class prototype $\bf{p}^\prime_n$ composed by the prompt and the pre-trained normal prototype.
\begin{equation}
\label{eq:fine_tune}
{L_{{{pt }}}} = \sum\limits_{i = 1}^{\left| {{\mathcal V}_l^{Test}} \right|} \left\| {{{\bf{r}}_i} - {\bf{p}^\prime_n}} \right\|_2^2,
\end{equation}
where the residual feature ${\bf r}_i$ is fixed since the learnable parameters in GNN are frozen while ${\bf p}^\prime_n$ is adjusted based on the newly added trainable parameters.

\subsection{GAD Using AnomalyGFM}
\noindent\textbf{Zero-shot Inference.}
To build a foundation model for GAD with zero-shot learning capabilities, we directly apply AnomalyGFM pre-trained on auxiliary graphs to the test graph $\mathcal{G}_{test}$, meaning that $\Theta_n$ and $\Theta_a$  in $\Phi({\bf{z}}_n; \Theta_n)$ and  $\Phi({\bf{z}}_a; \Theta_a)$ and the trainable parameters $\bf{W}$ in GNN are all frozen. This process allows us to obtain the residual features for each node in the test graph, the graph-agonistic normal prototype ${\bf{p}}_n$ and abnormal prototype ${\bf{p}}_a$. The anomaly score is determined by computing the similarity of the target node's representations residual with the two prototypes ${\bf{p}}_a$ and ${\bf{p}}_n$.
% respectively providing a consistently identical way for abnormality measurements. The 
Formally, the anomaly score $s_i$ for the node $v_i$ is defined as,
\begin{equation}
\label{eq:scoring}
{{s_i} =  \exp \left( {{\bf{r}}_i^{\rm{T}}{{\bf{p}}_{a}}} \right) + \beta \exp \left( -{{\bf{r}}_i^{\rm{T}}{{\bf{p}}_n}} \right)},
\end{equation}
where $\beta$ is the hyperparameter that controls the weight of two parts. Note that the anomaly scoring calculation consists of the direct similarity with ${\bf p}_a$  and the inverse similarity with ${\bf p}_n$. 

\noindent\textbf{Few-shot Inference.}
In the few-shot inference,  the original parameters in GNN, mapping functions, and newly added parameters for refining the normal class prototype are frozen. 
The anomaly score of ${{\mathcal V}_u^{Test}}$ is calculated based on the prompt-tuning-based updated normal class prototype $\bf{p}_n^\prime$ and the pre-trained abnormal class prototype ${\bf{p}}_{a}$, which is defined as:
\begin{equation}
\label{eq:scoring_few}
{{s_i} =  \exp \left( {{\bf{r}}_i^{\rm{T}}{{\bf{p}}_{a}}} \right) + \beta \exp \left( -{{\bf{r}}_i^{\rm{T}}\bf{p}_n^\prime} \right)}. 
\end{equation}

\noindent\textbf{Inference on Very Large-scale Graphs.}
Most existing GAD methods typically require loading all nodes of the graph, which often leads to poor scalability on large graphs during inference. By learning the graph-agnostic prototypes, AnomalyGFM can generalize to very large-scale graphs through a subgraph-based inference approach. AnonalyGFM can effectively infer the anomaly score without considering the entire graph structure, eliminating the bottleneck of loading the full graph for GAD inference. This subgraph inference is also desired in privacy-sensitive settings where we do not want to reveal the entire graph structure to the detection models.
% which fully addresses issues such as computational efficiency and privacy protection in GAD. 
Specifically, AnomalyGFM operates in a way that given a test node $v_i$ and its subgraph $\mathcal{S}\left(v_i\right)$ with size $s$ are given, it first computes its residual feature ${\bf{r}}_i$ as:
\begin{equation}
\label{eq:residual_subgraph}
    {
\mathbf{r}_i=\mathbf{h}_i-\frac{1}{\left|\mathcal{S}\left(v_i\right)\right|} \sum_{v_j \in \mathcal{S}\left(v_i\right)} \mathbf{h}_j
    }\, ,
\end{equation}
where ${\mathcal{S}}\left(v_i\right)$ is extracted using a random walk starting with the target node $v_i$. Then, AnomalyGFM infers the anomaly score using the similarity of the residual feature to the normal and abnormal class prototypes as defined in Eq. (\ref{eq:scoring}).

\section{Experiments}

\subsection{Experimental Setup}

\begin{table}[!t]
\centering
\setlength{\tabcolsep}{0.50mm}
\caption{Key statistics of GAD datasets used in our experiments. Ano. indicates the number (rate) of anomalies in the graph and Sim. indicates the global average edge similarity. 
% \gs{Tolokers has more than 20\% of nodes being anomalies, which is questionable for an AD dataset.}
}
\scalebox{0.80}{
\begin{tabular}{llccccccc}
\toprule
\hline
\textbf{Dataset} &\textbf{Domain} &\textbf{\#Nodes} & \textbf{\#Edges} &\textbf{\#Feat.} & \textbf{Ano.}  & \textbf{Sim.} \\
\hline
Facebook  & Social Networks& 1,081 &55,104 &576 &25(2.31\%) & 0.690\\
% Weibo & Social Networks &8,405 &407,963 &400 &868(10.30\%)\\
Reddit  & Social Networks &10,984 &168,016 &64 &366(3.33\%) &0.997\\
Amazon   & Co-review   &10,244 &175,608  &25 & 693(6.66\%) & 0.645\\
Disney &Co-purchase & 124 &335  & 28   & 6(4.8\%)   & 0.804\\
% YelpChi  & Co-review  &24,741 &49,315 &32 &1,217(4.91\%) &0.915\\
Amazon-all & Co-review    & 11,944 &  4,398,392&25 & 821(6.87\%)& 0.645\\
YelpChi-all  & Co-review &45,941 &3,846,979 &32 &6,674(14.52\%) & 0.905 \\
Tolokers & Work Collaboration & 11,758 &519,000 &10 & 2,566(21.8\%)  & 0.814\\ 
Question & Social Networks  &48,921 &153,540 &301 &1,460(2.98\%) &0.679\\ 
Elliptic & Bitcoin Transaction &46,564& 73,248 &93 &4,545 (9.8\%) & 0.356 \\
T-Finance & Transaction Record &39,357 &21,222,543 &10 &1,803(4.6\%) &0.107 \\
T-Social & Social Friendship &  5,781,065&  73,105,508&  10&  174,280(3.0\%) & 0.307\\
% BlogCatalog  & Social Networks & 5,196  &171,743  & 8,189 & 300(5.8\%) \\
% ACM  & Citation Networks & 16,484 &  71,980 & 8,337  &597(3.6\%)   \\
% Flickr  & Social Networks & 7,575  & 239,738  & 12,407      &450(5.2\%)   \\
\midrule
\bottomrule
\end{tabular}
}
\label{tab:dataset}
\end{table}

\noindent \textbf{Datasets.} We utilize 11 benchmark datasets from social networks, finance, and co-review domains for evaluation. These datasets are summarized in Table \ref{tab:dataset}, with additional details provided in Appendix \ref{app:dataset}. The social network datasets consist of Facebook \cite{xu2022contrastive}, Reddit \cite{kumar2019predicting}, Question \cite{platonov2023critical} and T-Social \cite{tang2022rethinking}. The financial datasets include T-Finance \cite{tang2022rethinking} and Elliptic \cite{weber2019anti}. For the co-review category, we use Amazon \cite{dou2020enhancing}, Amazon-all \cite{dou2020enhancing}, and YelpChi-all \cite{dou2020enhancing}. Additionally, the Disney \cite{sanchez2013statistical} and Tolokers \cite{platonov2023critical} datasets, derived from co-purchase and work collaboration networks, are also included. 

\begin{table*}[t]
\caption{AUROC and AUPRC results for zero-shot GAD on nine real-world datasets, with the models trained on Facebook only. For each dataset, the best performance per column within each metric is boldfaced, with the second-best underlined. Avg. denotes the average performance. P-values are from one-tailed Wilcoxon signed rank tests.
% \gs{pls perform one-tailed Wilcoxon signed rank test for AnomalyGFM against each of the competing methods, and add the p-value into an additional right-hand column: will need to pair AnomalyGFM with one competing method, and do it iteratively for all competing methods. Do the tests separately for AUROC and AUPRC results, meaning we have 11 pairs for AUROC-based significance test, and 11 pairs for AUPRC-based test. Pls do the same for the comparison in tables 3 and 4.}
% All results are the average of three times runs with different random seeds. 
}
\label{tab:main}
\begin{center}
\setlength{\tabcolsep}{1.50mm}
\scalebox{0.84}{
\begin{tabular}{c|c|ccccccccc|cc}
\hline
\hline
\multirow{2}*{\textbf{Metric}}&\multirow{2}*{\textbf{Method}} & \multicolumn{9}{c|}{\textbf{Dataset}} &\multirow{2}*{\textbf{Avg.}} &\multirow{2}*{\textbf{p-value}}\\
& &Reddit  &Amazon & Disney &Amazon-all &YelpChi-all &Tolokers & Question &Elliptic &T-Finance&&\\
\hline
\multirow{15}{*}{AUROC}
% \rowcolor{lightgray} 
&\multicolumn{12}{c}{Unsupervised Methods}\\
\cline{2-13}
&AnomalyDAE (ICASSP'20)   &0.5016 &0.5818   &0.4853  &0.7228&0.5002&\underline{0.5948} &0.4311&0.4197 &0.2324 &0.4966&0.007\\
& CoLA (TNNLS'21)     &0.4623 &0.4580 &0.4696&0.4091&0.4879  &0.4501&0.4945&0.5572 &0.4889 &0.4752&0.003 \\
% &HCM-A (ECML-PKDD'22) &0.5387  &0.4784 &0.5000&0.5056&0.5023&0.5172&\first{0.5731}&0.4160&0.2975&0.2014\\
&TAM (NeurIPS'23)      &\underline{0.5725} &0.4720  &0.4773 &0.7543 &0.4216&0.5351&0.5119&0.3282 &0.2990 &0.4857&0.003\\
&GADAM (ICLR'24)   &0.4532&0.6646  &0.4288 &0.5959&0.4829 &0.4832&\textbf{0.5594}&0.3922& 0.1382 &0.4664&0.007\\
\cline{2-13}
&\multicolumn{12}{c}{Supervised Methods}\\
\cline{2-13}
&GCN (ICLR'17)   &0.5645  &0.5988  &0.5000 &0.7195&0.5486 &0.5319&0.5161&\textbf{0.7640} &0.2345 &0.5531&0.039\\
&GAT (ICLR'18)    &0.5000  &0.4981  &0.5175 &0.5005&0.4802 &0.5030&0.4577&\underline{0.6588} &0.5072 &0.5136&0.007\\
&BWGNN (ICML'22) &0.5208  &0.4769  &0.6073 &0.3648 &0.5282&0.4877&0.4404 & 0.5843&  \underline{0.5457} &0.5062&0.003\\
&GHRN (WebConf'23)   &0.5253 &0.4560  &0.5336 &0.3382  &0.5125&0.4860 & 0.4535& 0.5400&0.5324  &0.4863&0.003\\
&XGBGraph (NeurIPS'23) &0.4601  &0.4179 &\underline{0.6692} &0.7950 &0.4945 &0.5462&0.5095&0.4274&0.3402 &0.5177&0.003\\
\cline{2-13}
&\multicolumn{12}{c}{Generalist Methods}\\
\cline{2-13}
&GraphPrompt (WebConf'23)  & 0.4677 &0.4904 &  0.5192& 0.3215 &0.4976&0.4779  &0.4204  &0.3221  & 0.5405 &0.4508&0.003\\
&UNPrompt (IJCAI'25)   &0.5337   &\underline{0.7525} & 0.6412 &\underline{0.7962} & \underline{0.5558} &\textbf{0.6853}&0.4757&0.5901& 0.2318 &\underline{0.5847}&0.074\\
% &GraphPrompt \\
% &All in one\\
\cline{2-13}
&AnomalyGFM   & \textbf{0.5974} &\textbf{0.8417} &\textbf{0.6751} &\textbf{0.9032} &\textbf{0.5791} &0.5843  &\underline{0.5280}
 & 0.6195 &   \textbf{0.5614}  &\textbf{0.6544}&/ \\

\hline

\multirow{15}{*}{AUPRC}
&\multicolumn{12}{c}{Unsupervised Methods}\\
\cline{2-13}
&AnomalyDAE (ICASSP'20)  &0.0327&0.0833  &0.0566 &0.1921 &0.1484 &0.1876 &0.0241&0.0798&0.0274  &0.0924&0.003\\
&CoLA (TNNLS'21)     &0.0391&0.0669 &0.0701  &0.0861 &0.1466 &0.0848&0.0292&0.0998& 0.0430&0.0739&0.007\\
% &HCM-A (ECML-PKDD'22)    &0.0391 &0.0669 &0.0511&0.0861 &0.1466&0.0848& \first{0.0428}&0.0383&0.0776&0.0355\\
&TAM (NeurIPS'23)      &\underline{0.0413} &0.0666 &0.0628 &{0.1736} &0.1240 &0.0970&0.0307&0.0697 &0.0332 &0.0776&0.007\\
&GADAM (ICLR'24)  &0.0293 &0.1562  &0.0651 & 0.1595& 0.1371 &0.1001&\underline{0.0395}&0.0733 &0.0261 &0.0873&0.003\\
\cline{2-13}
&\multicolumn{12}{c}{Supervised Methods}\\
\cline{2-13}
&GCN (ICLR'17)      &\textbf{0.0439} &0.0891  &0.0484 &0.1536  &0.1735 &0.1060&0.0387&\textbf{0.1963}& 0.0274&0.0974&0.074\\
&GAT (ICLR'18)       &0.0329  &0.0688 &0.0530 &0.0696 &0.1400 &0.0822&0.0259&0.1366 &0.0463 &0.0728&0.003\\
&BWGNN (ICML'22)  &0.0389  &0.0652  &0.0624&0.0586 &0.1605 &0.1030 &0.0257 &0.1158 &0.0479 &0.0753&0.007 \\
&GHRN (WebConf'23)    &0.0407 &0.0633 &0.0519 &0.0569 &0.1505  &0.0957 &0.0259 & 0.1148 &0.0457 &0.0717&0.007 \\
&XGBGraph (NeurIPS'23)   & 0.0330 &0.0536  & 0.1215 &0.2307 &0.1449&0.1256&0.0306&0.0816&\textbf{0.0754}&0.0996&0.027\\
\cline{2-13}
&\multicolumn{12}{c}{Generalist Methods}\\
\cline{2-13}
&GraphPrompt  (WebConf'23) &0.0334 &0.0661 &0.0610 &0.0666 &0.1542 &0.2070 &0.0266 & 0.0664&  0.0492&0.0811&0.003\\
&UNPrompt (IJCAI'25)   &0.0351  &\underline{0.1602} &\underline{0.1236} &\underline{0.2430} &\underline{0.1810} &\underline{0.2219} &0.0348&0.1278& 0.0279 &\underline{0.1283}&0.003\\
\cline{2-13}
&AnomalyGFM  & 0.0387  &\textbf{0.5790}  & \textbf{0.1242} &\textbf{0.6820 } & \textbf{0.1819} & \textbf{0.2749
} &\textbf{0.0397} &\underline{0.1371}  & \underline{0.0593} &\textbf{0.2352}&/   \\
\hline 
\hline 
\end{tabular}
}
\end{center}
\end{table*}

\noindent \textbf{Zero-shot Competing Methods.} For the zero-shot setting,  we evaluate AnomalyGFM against the state-of-the-art approaches from three main categories as follows (1) unsupervised methods: AnomalyDAE \cite{fan2020anomalydae}, CoLA \cite{liu2021anomaly}, TAM \cite{qiao2024truncated}, and GADAM \cite{chen2024boosting}; (2) supervised methods:  GCN \cite{kipf2016semi}, GAT \cite{velivckovic2017graph}, BWGNN \cite{tang2022rethinking}, GHRN \cite{gao2023addressing}, and XGBGraph \cite{tang2023gadbench}; and (3) general GFM methods, including GraphPrompt \cite{liu2023graphprompt} for general graph tasks, and zero-shot GAD UNPrompt \cite{niu2024zero}. 
% The detailed descriptions of each zero-shot competing method are provided in Appendix \ref{app:competingmethods}.

\noindent \textbf{Few-shot Competing Methods.} 
In few-shot setting, three generalist methods are icluded, \ie GPPT \cite{sun2022gppt}, GraphPrompt \cite{liu2023graphprompt}
% , designed for node classification tasks, 
and ARC \cite{liu2024arc}.
% developed for few-shot graph anomaly detection.

% Detailed descriptions of each generalist competing method are provided in Appendix. \ref{app:competingmethods}. 

% In this setting, we assume that only a very limited number of normal labels are known, given that the overwhelming presence of normal samples in GAD and the equivalent labeling cost associated with random sampling \cite{qiao2024generative,liu2024arc}.  

% In this scenario, AnomalyGFM is evaluated against semi-supervised methods by adapting unsupervised methods to the semi-supervised setting. The comparison includes DOMINANT \cite{ding2019deep}, OCGNN \cite{wang2021one}, AEGIS \cite{ding2021inductive}, and GGAD \cite{qiao2024generative}, a method specifically designed for semi-supervised settings.

\noindent \textbf{Evaluation Metrics.} Following the previous studies \cite{liu2024arc,tang2022rethinking, qiao2024truncated}, two widely-used metrics, AUROC and AUPRC, are used to evaluate the performance of all methods. For both metrics, a higher
value denotes better performance. Moreover, for each method, we report the average performance for 3 independent runs with different random seeds. One-tailed Wilcoxon signed rank test is performed to evaluate the statistical significance of the performance.
% of AnomalyGFM against each of its competing methods.

\noindent \textbf{Implementation Details.}
For all competing methods and AnomalyGFM, the code is implemented with Pytorch 1.11
, PyG 2.1.0, DGL 1.0.1, and Python 3.8.19. All experiments are conducted on a Linux server with an Intel CPU (Intel Xeon Gold 6346 3.1GHz) and an Nvidia RTX3090 GPU with 24 GB GPU memory.
% In the experiments, we choose one small dataset Facebook as the training source and evaluate on the other datasets across different domains. 
The pre-train model is optimized using the Adam optimizer \cite{kingma2014adam} with 300
epochs and a learning rate of $1e-4$ by default. The dimension of hidden space for representation and the prototype size are uniformly set to 300.  The parameter $\alpha$ in the loss function is set to one by default. 
The setting of $\beta$ is determined based on the prior we gain from the global average similarity information provided in Table \ref{tab:dataset} (see Sec. \ref{app:scoring} for details).
The $\mu$ and $\sigma$ of the Gaussian distribution for both normal and abnormal class prototypes are set to zero and one by default. 
% See Appendix~\ref{app:sensitive}, for the performance of AnomnalyGFM under different initialization hyperparameters. 
The size of the subgraph in subgraph inference is set to five by default. The implementation of competing methods is based on their publicly available official source code.

\begin{table*}[t]
\caption{Results on the nine real-world GAD datasets under \underline{1/5/10-shot} with models tuned on the few-shot data.}
 % For each dataset, the best performance per column within each metric is boldfaced, with the second-best underlined.
% Avg. denotes the average performance.}
% For each dataset, the best performance per column within each metric is boldfaced, with the second-best underlined. All results are the average of three times runs with different random seeds.
 % \gs{how about giving WWup the 100-shot experiments and focusing on the 10-shot experiments only at this stage? Evaluate and add also the results of DOMINANT, AEGIS, TAM, and GGAD under the 10-shot setting.} \gs{can we add the more recent method GADAM in this comparison?}
\label{tab:few_shot}
\begin{center}
\setlength{\tabcolsep}{0.65mm}
\scalebox{0.9}{
\begin{tabular}{c|c|c|ccccccccc|cc}
\hline
\hline
\multirow{2}*{\textbf{Metric}}&\multirow{2}*{\textbf{Setting}} &\multirow{2}*{\textbf{Method}} &    \multicolumn{9}{c|}{\textbf{Dataset}} &\multirow{2}*{\textbf{Avg.}} &\multirow{2}*{\textbf{p-value}}\\
&  & &Reddit  & Amazon & Disney &Amazon-all &YelpChi-all &Tolokers & Question &Elliptic &  T-Finance && \\
\hline

\multirow{9}{*}{AUROC} & \multirow{4}{*}{1-shot}
&GPPT (KDD'22) &\underline{0.5000} &\underline{0.5303} &\underline{0.4997} &\underline{0.5010} &\underline{0.5000} &0.5061 &0.4921 &\underline{0.6162} &0.3647 &\underline{0.5011}  & 0.003 \\
 &&GraphPrompt (WebConf'23)&0.4216&0.4882&0.4223&0.2631&0.4811&\underline{0.5328}&0.4086&0.6001& \underline{0.4000} &0.4464& 0.004\\ 
& &ARC (NeurIPS'24)  &0.4899  &0.4571  &  0.3578 &0.4570  &\underline{0.4910}  & 0.4667 &\textbf{0.5865} &0.2904  &0.2484  &0.4272& 0.008 \\
& &AnomalyGFM & \textbf{0.5922}  &\textbf{0.8531}  & \textbf{0.6649}   &\textbf{0.8972}  & \textbf{0.5872}  & \textbf{0.5898} & \underline{0.5303}  &  \textbf{0.6199} & \textbf{0.5916}&\textbf{0.6584}& /\\
\cline{2-14}
& \multirow{4}{*}{5-shot}
&GPPT (KDD'22) &\underline{0.5000} &\underline{0.5098} &0.5000 &\underline{0.5051} &0.5000 &0.5181 &0.4959 &0.5736 &0.2609 &0.4848  &0.003\\
&&GraphPrompt (WebConf'23) &0.4406&0.4900& \underline{0.6497}&0.4726& \underline{0.5359}&\underline{0.5381}&0.4069&\underline{0.6012}&\underline{0.4069} &\underline{0.5046}&0.003\\ 
& &ARC (NeurIPS'24)  &\underline{0.4720}  & 0.4458 &  0.4435 & 0.4473   & 0.5112  &0.4746 &\textbf{0.5906}  & 0.2714  &  0.2168 &0.4303&0.007 \\
& &AnomalyGFM & \textbf{0.6023 }  & \textbf{0.8600} & \textbf{0.6613} & \textbf{0.9011}   & \textbf{0.5951} &\textbf{0.6095}  &\underline{0.5426} &\textbf{0.6119}   & \textbf{0.6248}&\textbf{0.6676}& / \\
\cline{2-14}

 & \multirow{4}{*}{10-shot}
&GPPT (KDD'22) & \underline{0.5000} &  \underline{0.5087}  &  0.4769&  0.5023&  0.5000&  0.4971&  0.5047&  0.4212&  0.5539& 0.4961  &0.003\\
& &GraphPrompt (WebConf'23) &0.4321 &0.4906 &\underline{0.6314} &\underline{0.7167} &\underline{0.5367}&\underline{0.5329} &0.3826&\underline{0.6221}& \underline{0.4260} &\underline{0.5301}&0.007\\ 
& &ARC (NeurIPS'24)  & 0.4867  & 0.4323  & 0.4769 & 0.4467  & 0.5145  & 0.4786  & \textbf{0.5901}  &  0.2644&0.2298  &0.4355&0.003\\
& &AnomalyGFM & \textbf{0.6252}   &\textbf{0.8649}  &  \textbf{0.6649} &\textbf{0.9215}  & \textbf{0.6064} & \textbf{0.6140}  & \underline{0.5611} & \textbf{0.6303}  & \textbf{0.6283} &\textbf{0.6796}& /\\
 
\hline
\hline
\multirow{9}{*}{AUPRC} & \multirow{3}{*}{1-shot}
 &GPPT (KDD'22) &\underline{0.0333} &\underline{0.0766} &\underline{0.0488} &\underline{0.0687} &\underline{0.1453} &0.2204 &0.0294 &\underline{0.1239} &\underline{0.0432} &\underline{0.0877} &0.003\\
&&GraphPrompt (WebConf'23) &0.0283&0.0680&0.0486&0.0426&0.1113&\underline{0.2321}&\textbf{0.0448}&0.1108& 0.0302&0.0796&0.012\\ 
& &ARC (NeurIPS'24)   &0.0332  &  0.0581  & 0.0453 & 0.0590  & 0.1402  & 0.2122 &\underline{0.0468}  & 0.0701 &  0.0277 &0.0769&0.011  \\
& &AnomalyGFM  & \textbf{0.0398}   &\textbf{0.5801}  & \textbf{0.1223}  & \textbf{0.6921} &\textbf{0.1852} & \textbf{0.2786}   &  0.0332  &  \textbf{0.1401}  &\textbf{0.0601}&\textbf{0.2368}& / \\

\cline{2-14}
& \multirow{3}{*}{5-shot}
 &GPPT (KDD'22) &\underline{0.0333} &\underline{0.0692} &0.0504 &\underline{0.0716} &0.1453 &0.2265 &0.0297 &0.1127 &\underline{0.0365} &0.0861 & 0.004\\
 &&GraphPrompt (WebConf'23) &0.0285&0.0681& \underline{0.0892}&0.0600&\underline{0.1661}&\textbf{0.2957}&0.0327&\underline{0.1416}&0.0360 &\underline{0.1019}& 0.009\\ 
& &ARC (NeurIPS'24)  &0.0312 &0.0571 &0.0546  & 0.0572  & 0.1464  & 0.2150 & \textbf{0.0471}  &0.0726  & 0.0267  & 0.0786& 0.011\\
& &AnomalyGFM   & \textbf{0.0401} & \textbf{0.5831} & \textbf{0.1257} &\textbf{0.6985} & \textbf{0.1918} &\underline{0.2866}  &\underline{0.0336}  &  \textbf{0.1437} &\textbf{0.0622} &\textbf{0.2405}& /\\
\cline{2-14}
& \multirow{3}{*}{10-shot}
&GPPT (KDD'22) &\underline{0.0334}  & \underline{0.0691} & 0.0526 & 0.0698 &0.1453 &0.2178 &0.0301 &0.0905 & \underline{0.0511} &0.0844 &0.004\\
&&GraphPrompt (WebConf'23) &0.0278 &0.0681 &\underline{0.0848} &\underline{0.1427}&\underline{0.1649}&\textbf{0.2922}&0.0263&\underline{0.1421}&0.0382 &\underline{0.1096}& 0.007\\ 
& &ARC (NeurIPS'24)  & 0.0327   &  0.0557  &  0.0743  & 0.0583 & 0.1491 & 0.2168  &\textbf{0.0463}  &  0.0677   & 0.0272  &0.0809& 0.011\\
& &AnomalyGFM  & \textbf{0.0444}   & \textbf{0.5895} &  \textbf{0.1399} & \textbf{0.7124}  & \textbf{0.1990}  & \underline{0.2897}  & \underline{0.0346}  & \textbf{0.1570} &   \textbf{0.0644} &\textbf{0.2478}& /\\
\hline
\hline
\end{tabular}
}
\end{center}
\end{table*}

\subsection{Zero-shot GAD Performance}
The AUROC and AUPRC results under the zero-shot setting are shown in Table \ref{tab:main} where the model is trained on Facebook and inference is made on other datasets without any further tuning. Similar results can be found when the models are trained on the Amazon dataset in Appendix \ref{app:train_amazon}. From the table, we observe
(1) AnomalyGFM consistently outperforms the three types of competing methods, including unsupervised methods, supervised methods, and generalist methods on most datasets. Specifically, AnomalyGFM achieves the best AUROC performance on six datasets and the best AUPRC performance on the five datasets, with up to 11\% improvement in AUROC and 44\% improvement in AUPRC when compared to the best-competing method. The main reason is that AnomalyGFM performs inference based on two discriminative graph-agonistic prototypes distilled from the node residual feature, supporting strong generalization across the datasets. 
(2) Unsupervised methods often fail to produce consistently good results across the datasets due to the fact that each of their model designs is devised in a way to focus on some specific abnormal patterns for the training graphs, making them struggle to generalize to the new graphs.
(3)  Most supervised methods perform well in zero-shot inference, particularly simple approaches like GCN, which achieves top performance on the Elliptic dataset. This can be attributed to their simple model design, avoiding heavy overfitting to the training graph. However, the performance still underperforms AnomalyGFM, demonstrating that the node representation residual is more generalized and our prototype alignment module provides a strong add-on to the binary loss function in our method.
% , as the node representations are mapped to the unified feature space while maintaining the anomaly discriminability.  
(4) Generalist methods, such as GraphPrompt, which are designed for general node/graph classification, struggle to perform well on GAD. This is due to the irregular nature of the abnormalities in different GAD datasets, hindering the applicability of the regular patterns/knowledge these generalist method learned from other tasks to GAD. AnomalyGFM outperforms UNPrompt on all datasets except Tolokers, which can be attributed to the node representation residual-based class prototypes offers more generalization power than the latent attribute predictability used in UNPrompt. 
% Additionally, we simultaneously use measurements from both the normal and abnormal prototypes compared to relying solely on node affinity.
Further, our statistical significance test results show that the superiority of AnomalyGFM over all competing methods is significant at the 90\%/95\% confidence level in both metrics.

\subsection{Few-shot GAD Performance}
To demonstrate the effectiveness in the few-shot scenarios, the comparison of AnomalyGFM to GraphPrompt \cite{liu2023graphprompt} and ARC \cite{liu2024arc} is done under 1/5/10-shot. The AUROC and AUPRC results are shown in Table \ref{tab:few_shot}. In the few-shot GAD scenario, AnomalyGFM outperforms these two generalist methods on most datasets in both AUROC and AUPRC. GraphPrompt continues to underperform AnomalyGFM since the presence of only one class of samples renders its fine-tuning ineffective, leading to suboptimal performance. ARC also leverages residual features 
\begin{table}[t]
\caption{Subgraph inference on the very large-scale graphs. ‘/’ indicates that the model cannot handle the dataset.}
\label{tab:subgraph}
\begin{center}
\setlength{\tabcolsep}{1.30mm}
\resizebox{0.4\textwidth}{!}{
\begin{tabular}{c|c|cc}
\toprule
\hline
\multirow{2}*{\textbf{Metric}} &\multirow{2}*{\textbf{Method}} & \multicolumn{2}{c}{\textbf{Dataset}}\\
 & &    T-Finance& T-Social\\
% \midrule
\hline
\multirow{8}*{AUROC}&\multicolumn{3}{c}{Unsupervised Methods}\\
\cline{2-4}
 & TAM (NeurIPS’23)   &0.2990 & / \\
 & GADAM (ICLR’24)   &0.1382 &\underline{0.5155}  \\
 \cline{2-4}
 &\multicolumn{3}{c}{Supervised Methods}\\
 \cline{2-4}
 &BWGNN (ICML'22)   &\underline{ 0.5457} & 0.4964 \\
 &GHRN (WebConf'23)  &  0.5324 &0.4934\\
& XGBGraph (NeurIPS’23) & 0.3402   &0.4602  \\
\cline{2-4}
  &AnomalyGFM   & \textbf{0.7852} & \textbf{0.5991}\\
\cline{2-3}

\hline
\multirow{8}*{AUPRC}&\multicolumn{3}{c}{Unsupervised Methods}\\
\cline{2-4}
 & TAM (NeurIPS’23)  &0.0332  &  /\\
 & GADAM (ICLR’24)   &0.0261 &0.0285  \\
 \cline{2-4}
&\multicolumn{3}{c}{Supervised Methods}\\
\cline{2-4}
 &BWGNN (ICML'22)    &  0.0479 &0.0301\\
&GHRN (WebConf'23)    &  0.0457 &0.0303\\
& XGBGraph (NeurIPS’23)  &\underline{0.0754}  &\underline{0.0305} \\
\cline{2-4}
 &AnomalyGFM  &\textbf{0.1059}  & \textbf{0.0398}\\
\cline{2-3}
\hline
\bottomrule
\end{tabular}}
\end{center}
\vspace{-1.5em}
\end{table}
and is specifically designed for few-shot GAD datasets achieving the best performance on the Question and the second performance on most datasets. Compared to ARC, AnomalyGFM achieves better sample-efficiency tuning due to the learning of class-level prototypes that are agnostic to different domains of graphs, enhancing the generalization across the graph, thereby yielding the best performance on the eight datasets. Tuning the abnormal prototype is also possible when a few labeled anomalies are available and the results are provided in App. \ref{app:tuning_abnormal}.

\subsection{Scale up to Very Large Graphs}
% AnomalyGFM can generalize to large-scale graphs through subgraph-based inference. It is also applicable when the entire graph structure and attribute are unknown. 
We choose two large-scale graphs T-Finance (with a large number of edges) and T-Social (with a large number of nodes)
to evaluate the generalization of AnomalyGFM on very large graphs. The results are shown in Table \ref{tab:subgraph}. We can observe that AnomalyGFM outperforms the unsupervised methods and supervised methods on these datasets. This is mainly because the learned data-independent prototypes can effectively measure the abnormality of node representation residual, even in the presence of subgraphs. The deviations of the target node from the contextual node are still preserved in the subgraph, thereby enabling effective subgraph-based inference.

\begin{figure}
\setlength{\abovecaptionskip}{0.2cm}
 \centering
 \includegraphics[width=2.75in,height=1.1in]{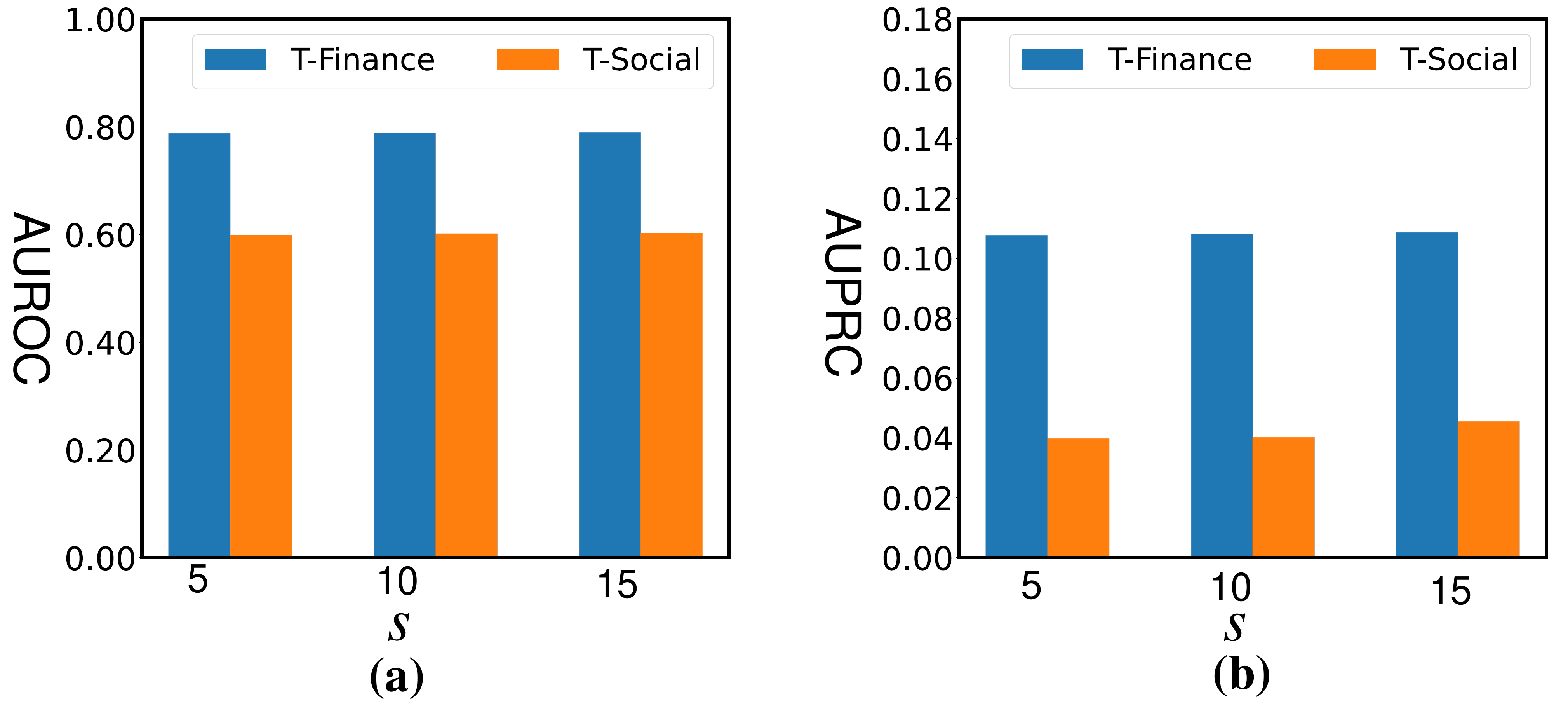 }\\
 \caption{AnomalyGFM performance w.r.t subgraph size $s$. 
}
 \label{fig:subgraph}
 \vspace{-1em}
\end{figure}

\begin{table*}[t]
\caption{Ablation study results for AnomalyGFM and its variants. }
% For each dataset, the best performance per column within each metric is boldfaced, with the second-best underlined. Avg. denotes the average performance.
\setlength{\tabcolsep}{1.0mm}
\label{tab:ablation}
\begin{center}
\scalebox{1.0}{
\begin{tabular}{c|c|cccccccccc}
\toprule
\hline
\multirow{2}*{\textbf{Metric}}&\multirow{2}*{\textbf{Method}} & \multicolumn{10}{c}{\textbf{Dataset}} \\
&  &Reddit & Amazon & Disney & Amazon-all  & YelpChi-all & Tolokers & Question & Elliptic &  T-Finance   &\textbf{Avg.}\\
% \midrule
\hline
\multirow{4}*{AUROC}

& FA  &\underline{0.5697}  &\underline{0.5680}  & 0.6073  & 0.5097  & 0.5211 &\underline{ 0.5594}  &\underline{0.5326}   &\textbf{0.7765} &\textbf{0.6674}   &\underline{0.5901}\\
&BCE    & 0.5445 & 0.4308 & \textbf{0.7542} & 0.5545  &\underline{0.5232}  &0.5187  & 0.4593  &  0.2581 &\underline{0.6581}  & 0.5223 \\
&BCE-R  & 0.5108  &  0.5314  & 0.4887   & \underline{0.5777}  & 0.4590    &0.4788  & 0.4601  & 0.5414  &0.5380   &0.5095 \\
&AnomalyGFM & \textbf{0.5974} &\textbf{0.8417} &\underline{0.6751} &\textbf{0.9032} &\textbf{0.5791} &\textbf{0.5843}  &\textbf{0.5380}
&\underline{ 0.6195} &0.5614  &\textbf{0.6555}\\
\cline{2-6}
\hline
\hline
\multirow{4}*{AUPRC}
&FA  &\textbf{0.0400} & \underline{0.0776}  & 0.0960   &  0.0684 &0.1555  &\underline{0.2478}  &\underline{0.0356} & \underline{0.1199}    &\textbf{0.1495} &\underline{0.1100}\\
&BCE    & \underline{0.0394} & 0.0544  &\textbf{0.2254}  & 0.0718&  \underline{0.1594}  &0.2264  & 0.0272  &  0.0287 &\underline{ 0.1389}  &0.1079\\
&BCE-R   & 0.0314 &0.0677  &  0.0685  &  \underline{0.0787} &  0.1294   & 0.2122 & 0.0305 &0.0658   & 0.0971   &0.0868\\
&AnomalyGFM  & 0.0387  &\textbf{0.5790}  & \underline{0.1242} &\textbf{0.6820 } & \textbf{0.1819} & \textbf{0.2749 } &\textbf{0.0397} & \textbf{0.1371} &0.0593  &\textbf{0.2352}\\
\cline{2-6}
\hline
\bottomrule
\end{tabular}}

\end{center}
\end{table*}
% \noindent \textbf{Performance of AnomalyGFM w.r.t. subgraph size.}
We also evaluate the sensitivity of AnomalyGFM with subgraph sizes and the results are shown in Fig. \ref{fig:subgraph}. The performance of AnomalyGFM remains stable with varying subgraph sizes, demonstrating its robust generalization capability in modeling large graphs. 

\subsection{Ablation Study}
To verify the effectiveness of each component of AnomalyGFM, we designed three variants including  (i) \textbf{BCE}
which replaces the similarity measurement with the predicted probability from the BCE loss function for anomaly scoring.
(ii) \textbf{BCE Residual (BCE-R)} which firstly removes the alignment loss and applies the BCE loss on the residual features to differentiate normal and abnormal nodes. Then, the predicted probability is used as the anomaly score.
(iii) \textbf{Feature Alignment (FA)} which replaces the alignment on the residual feature with the alignment on the original feature.
% All the results are shown in Table \ref{tab:ablation}. AnomalyGFM outperforms all the variants on most of the datasets indicating its effectiveness in capturing abnormality across graphs. 
The results are shown in Table \ref{tab:ablation}.
\textbf{FA} achieves the highest AUROC on the Elliptic and T-Finance datasets, suggesting that learning prompts from the original features can also be effective as the original feature also maintains anomaly discriminability but they aren't in the unified feature space leading to sub-optimal performance. \textbf{BCE} does not perform well except on the small dataset Disney.  It directly uses the predicted probability as the anomaly score, which makes it struggle to generalize to new graphs due to the differences between the training and test sets.
\textbf{BCE-R} does not achieve good performance compared to \textbf{BCE}. The reason is that although the residual features are discriminative, many samples located on the decision boundaries after being mapped to a unified feature space may not be effectively scored by the supervised classifier. In contrast, aligning the learnable prototype with the average node representation residual and using a similarity measurement for scoring can effectively discriminate the normal and abnormal nodes.

\subsection{Hyperparameter Sensitivity Analysis} \label{sec:sensitive}
We evaluate the sensitivity of AnomalyGFM w.r.t the size of the prototype $T$, hyperparameter $\alpha$, and commonality dimension $d^\prime$. Additional sensitivity analyses are in Appendix \ref{app:sensitive}.

\begin{figure}
\setlength{\abovecaptionskip}{0.2cm}
\centering
\includegraphics[width=2.65in,height=1.3in]{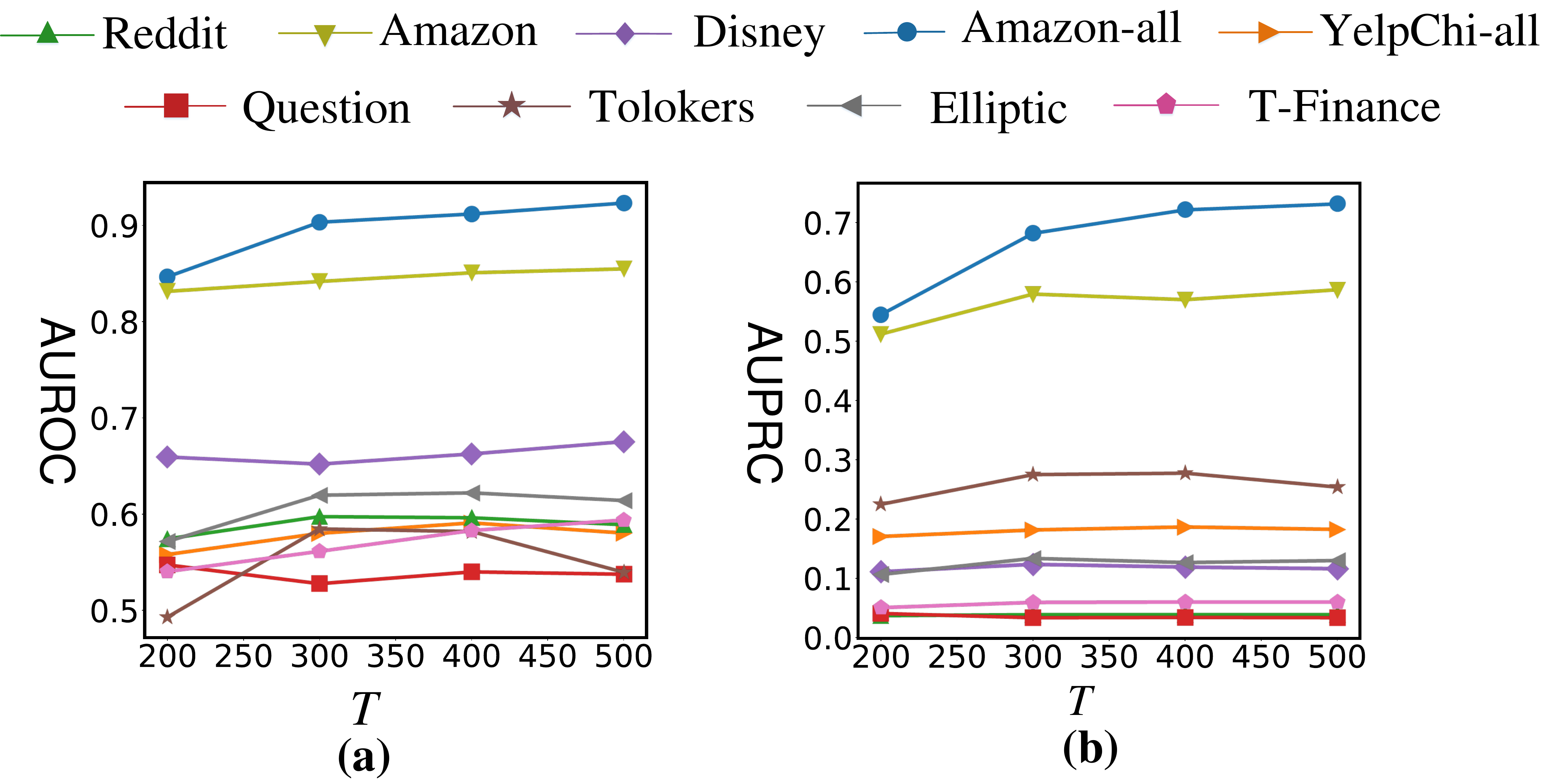 }\\
\caption{AnomalyGFM performance w.r.t prototype size $T$. 
}
\label{fig:hyper_size}
% \vspace{-1em}
\end{figure}

\noindent \textbf{AUROC and AUPRC results of AnomalyGFM w.r.t. prototype size $T$.} As shown in Fig. \ref{fig:hyper_size}, AnomalyGFM generally achieves better performance with increasing prototype size. This suggests that a larger prototype size provides more information and enhances the modeling capability of the foundation model.

\begin{figure}
\setlength{\abovecaptionskip}{0.2cm}
 \centering
 \includegraphics[width=2.85in,height=1.26in]{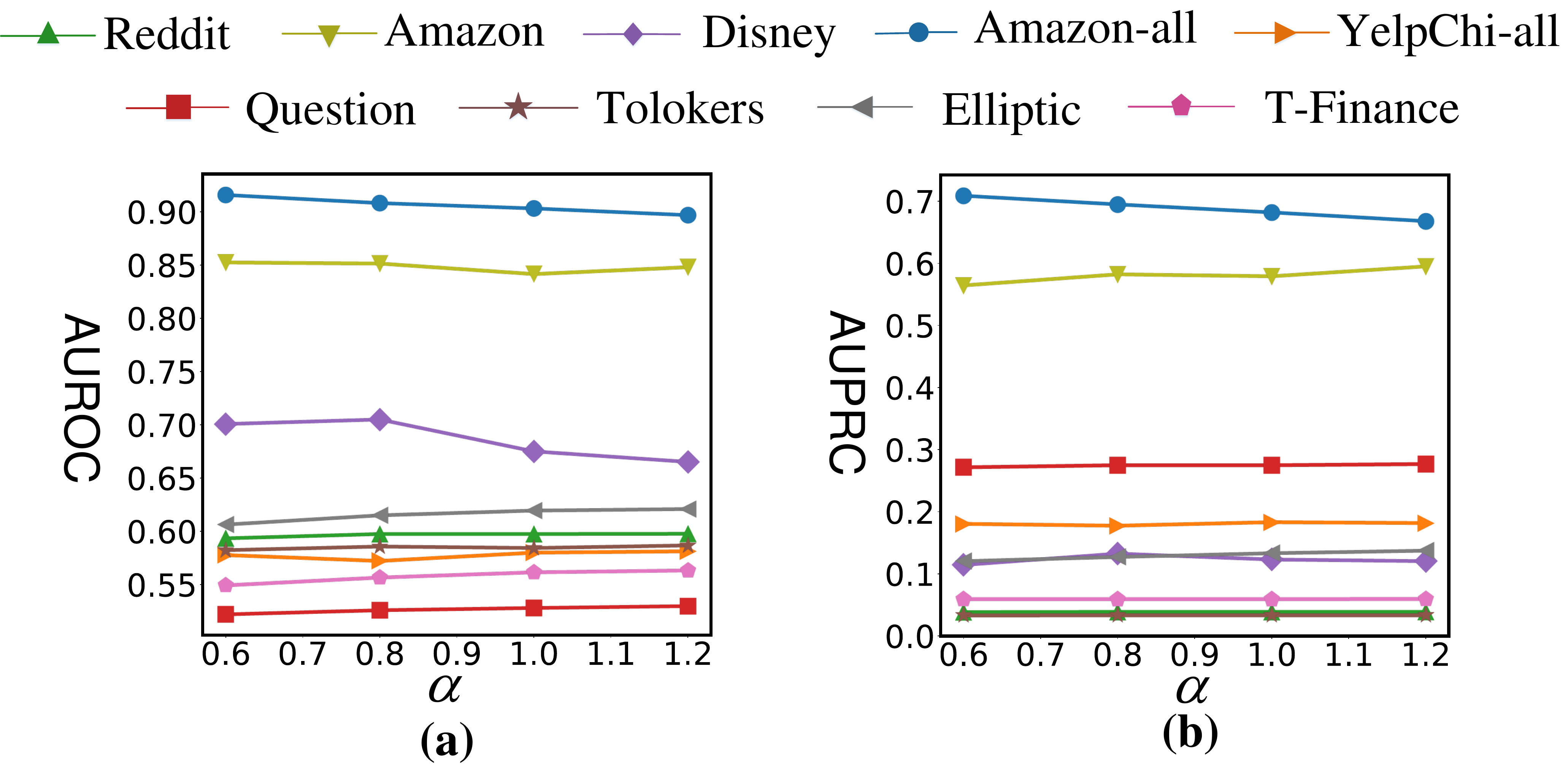 }\\
 \caption{ AnomalyGFM performance  w.r.t $\alpha$.}
 \label{fig:hyper_total}
  % \vspace{-1em}
\end{figure}

\noindent \textbf{AUROC and AUPRC results of AnomalyGFM w.r.t. $\alpha$.}
We vary the hyperparameter $\alpha$ in the range of [0.6, 1.2] with the interval of 0.1 and the results are shown in Fig. \ref{fig:hyper_total}. We can see that AnomalyGFM remains stable with different $\alpha$ on most datasets in terms of both metrics, demonstrating the robustness of AnomalyGFM against this hyperparameter.

\begin{figure}
\setlength{\abovecaptionskip}{0.2cm}
 \centering
 \includegraphics[width=3.15in,height=2.1in]{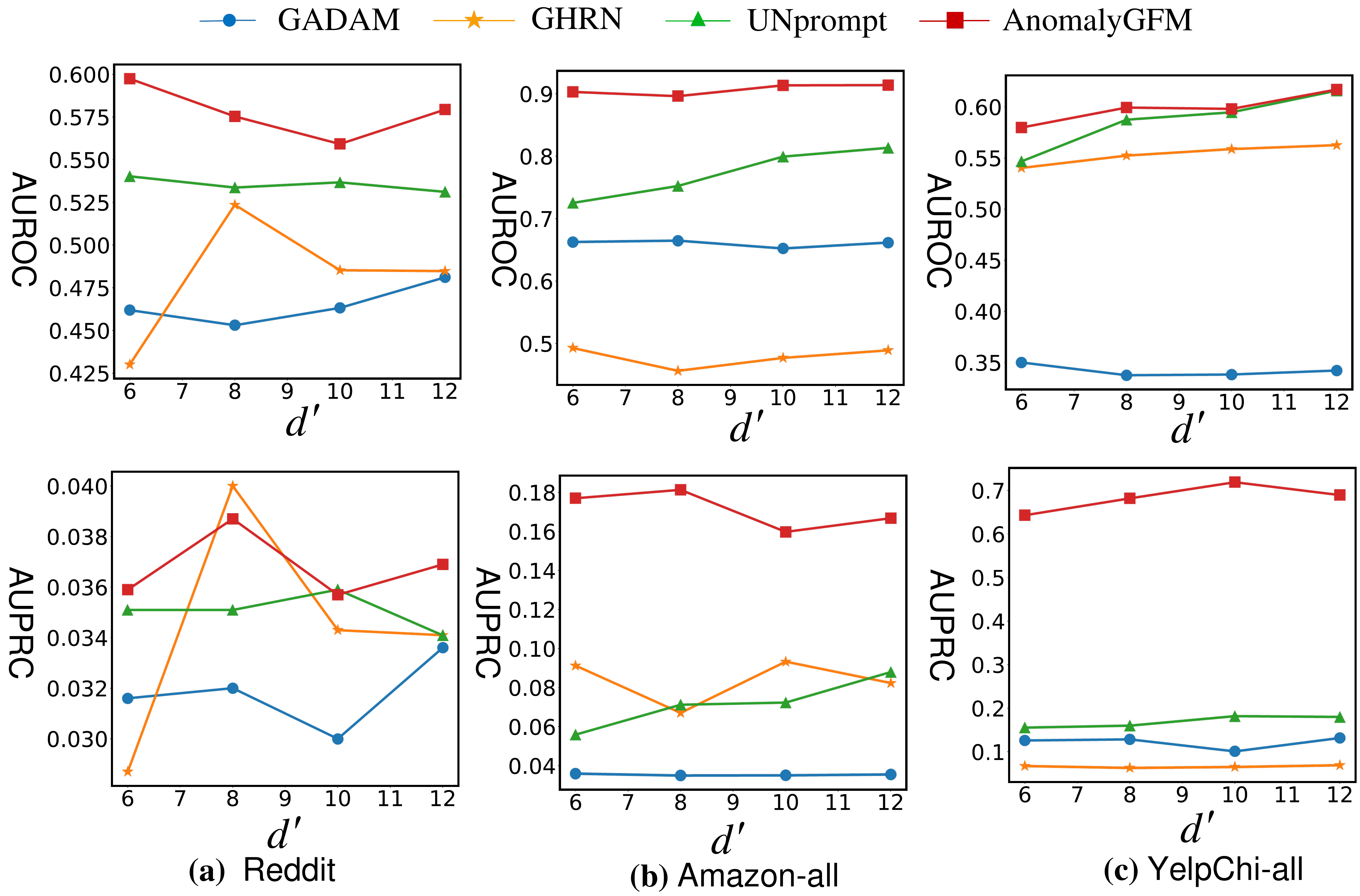 }\\
 \caption{ AUROC and AUPRC of AnomalyGFM w.r.t $d^\prime$ 
}
 \label{fig:dimension_roc}
% \vspace{-1em}
\end{figure}

\noindent \textbf{Sensitivity  of AnomalyGFM w.r.t common dimensionality $d^\prime$.} To further demonstrate AnomalyGFM's generalization, we evaluate AnomalyGFM
% with competing methods 
under different $d^\prime$. The results are shown in Fig. \ref{fig:dimension_roc}. AnomalyGFM outperforms all the competing methods under different dimensions.
% Specifically, we vary $d^\prime$ in the range of [6,12].
% indicating that AnomalyGFM is generalizable on the feature unification. 
The main reason is that the deviation between connected nodes is still preserved with different $d^\prime$ which can be efficiently measured by the graph-agnostic prototypes.
% The AUPRC results are provided in Appendix \ref{app:sensitive}.

\begin{figure}
\setlength{\abovecaptionskip}{0.2cm}
 \centering
 \includegraphics[width=2.85in,height=1.3in]{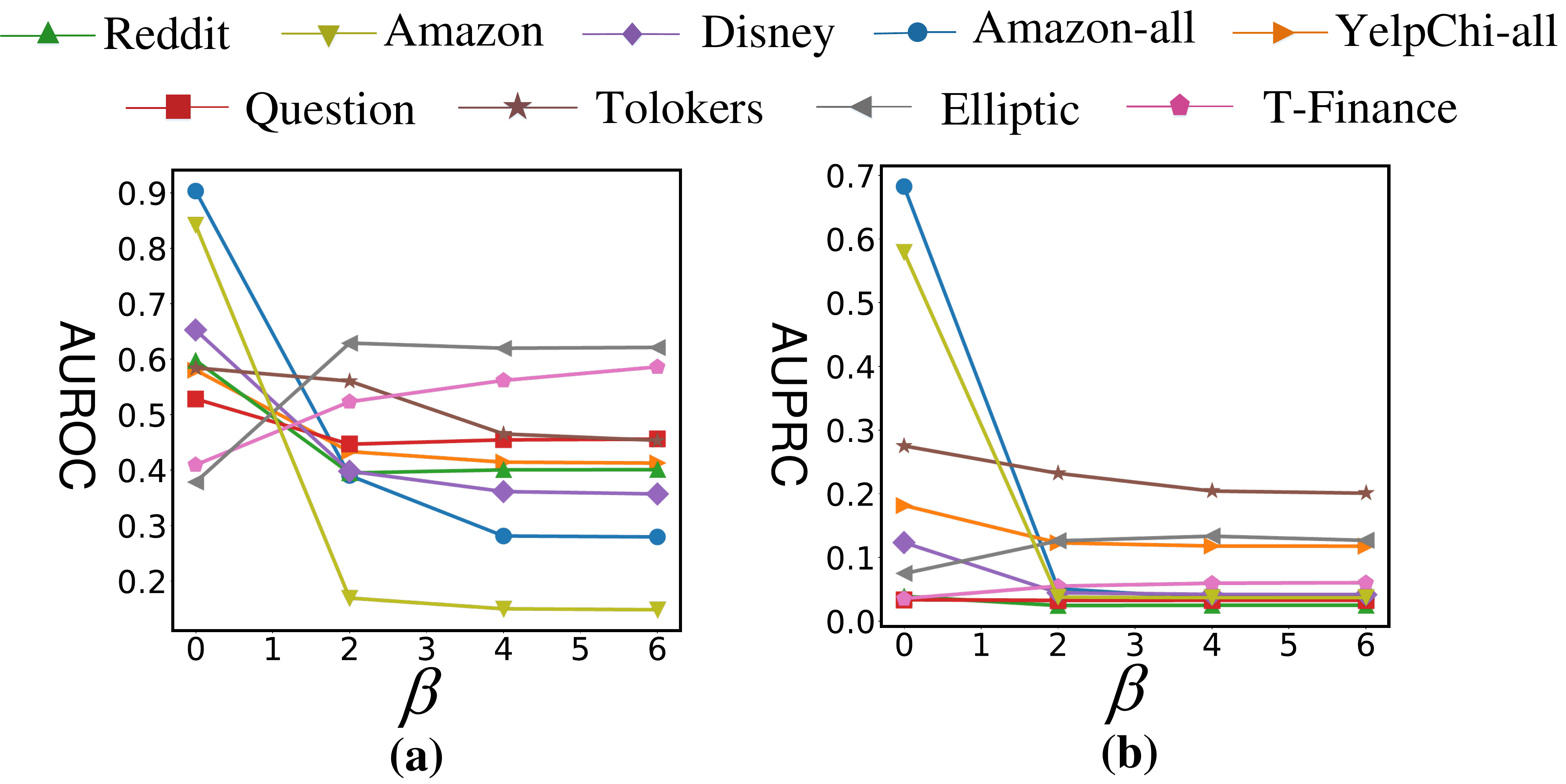 }\\
     \caption{ AUROC and AUPRC of AnomalyGFM w.r.t $\beta$. 
}
\label{fig:scoring}
\vspace{-1em}
\end{figure}

\noindent \textbf{Impact of $\beta$ in anomaly scoring.} 
\label{app:scoring}
The performance under different $\beta$ is shown in Fig. \ref{fig:scoring}. As shown in Table \ref{tab:scoring} in App. \ref{app:sensitive}, the set of $\beta$ values are determined based on the prior we gain from the global average similarity. Specifically, the hyperparameter in anomaly scoring  $\beta$ is set to zero by default for graphs with high global similarity such as Amazon and Yelp, and four for those with low global similarity such as Ellitipic and T-Finance, see the detailed global similarity information in Table \ref{tab:dataset}. In the few-shot scoring, $\beta$ is set to $0.5$ by default for graphs with high global similarity. The main reason for these settings is that for graphs with high global average similarity, where node features are relatively similar, the deviations in local neighborhood would be small, and the fluctuations in scoring based on the normal prototype may affect the overall anomaly score. Therefore, we solely use the abnormal class prototype or assign a smaller weight for the normal class prototype in scoring. For nodes with less similar features, we use a mix of normal and abnormal prototypes for scoring. In the few-shot setting, the labeled normal nodes help refine the normal prototypes, allowing the anomaly scores to be computed using both normal and abnormal prototypes.

\begin{table}[ht]
\caption{Training and inference time (seconds) comparison.}
\centering
\setlength{\tabcolsep}{0.35mm}
\begin{center}
\resizebox{0.48\textwidth}{!}{
\begin{tabular}{c|cccccc}
% \toprule
\hline
Methods&  AnomalyDAE & TAM & GAT & BWGNN & UNPrompt & AnomalyGFM\\
\hline
Training Time &86.04&479.70&2.43 &4.86 &2.08  &7.45 \\
Inference Time&170.46 &252.52  &91.88 &98.33&19.47 &11.00 \\
\hline
% \bottomrule
\end{tabular}}
\end{center}
\label{tab:running_time}
\vspace{-1em}
\end{table}
\section{Time Complexity Analysis}
In this section, we analyze the time complexity of AnomalyGFM. We build the GCN as the backbone of AnomalyGFM which takes $O(mdh)$, where $m$ is the number of non-zero elements in matrix $\bf A$ (the number of edge in graph), $d$ is the dimension of representation, and $h$ is the number of feature maps. The computation of the residual feature takes $O(N^2d)$. The MLP layer mapping the representation to the anomaly score takes $O(Nd)$. The feature mapping of normal class prototype and abnormal class prototype takes $O(2d^2)$. The alignment loss between
residual features and prototypes takes $O(2Nd)$. Thus, the overall complexity of the pre-training of AnomalyGFM is $O(mdh+N^2d+3Nd+2d^2)$. In the inference, we employ the residual feature's similarity with two prototypes which takes $O(2Nd)$. Therefore, the overall complexity of inference of AnomalyGFM is $O(mdh + N^2d+ 2d^2 +2Nd)$.
The complexity of inference on large-scale graph takes $O\left( {K\left( {mdh + {n^2}d + 2{d^2} + 2nd} \right)} \right)$
 with $n$ as subgraph size and $K$
 as the number of subgraphs. It is significantly lower than other methods, as they require loading entire graphs and performing reconstruction or affinity calculations, leading to higher space/time complexity. In Table~\ref{tab:running_time}, we report the running time comparison on Facebook and inference time on YelpChi-all.  AnomalyGFM requires much less training and inference time compared to TAM and AnomalyDAE, demonstrating the efficiency. For inference, AnomalyGFM is the most efficient method because scoring relies solely on measuring the similarity between node representation residuals and prototypes.

\section{Conclusion}
In this paper, we build a GAD-oriented graph foundation model, AnoamlyGFM, that can work effectively under both few-shot and zero-shot scenarios. AnomalyGFM is pre-trained to learn discriminative and data-independent prototypes by aligning them with the graph-agnostic node representation residuals. This provides a consistent and identical way for abnormality measurement using the similarity between residual node representation and the learned class prototypes, facilitating the strong generalization in both zero-shot and few-shot inference.
Extensive experiments on 11 datasets demonstrated the effectiveness and generalization of AnomalyGFM.  

\section*{Acknowledgments}
This research is supported by A*STAR under its MTC YIRG Grant (No. M24N8c0103), the Ministry of Education, Singapore under its Tier-1 Academic Research Fund (No. 24-SIS-SMU-008), the Lee Kong Chian Fellowship, and ARC under Grant DP240101349. 
% To enable our AnomalyGFM to generalize across various GAD tasks, including both graph-level and edge-level anomaly detection, we plan to incorporate these aspects into future work.

\bibliographystyle{ACM-Reference-Format}
\balance
\bibliography{sample-base}
\appendix
\section{Detailed Dataset Description} \label{app:dataset}
A detailed introduction of all the used datasets is given as follows.
\begin{itemize}
\item  Facebook  \cite{xu2022contrastive}: It is a social network where nodes represent users and edges indicate relationships between them. Anomalies are the nodes that either connect manually constructed clusters or display attributes differing from those of their contextual neighbors.

\item Reddit \cite{kumar2019predicting}: It is a user-subreddit graph, capturing one month’s worth of posts shared across various subreddits at Reddit. The users who have been banned by the platform are labeled anomalies. The text of each post is transformed into the feature vector and the features of the user and subreddits are the feature summation of the post they have posted.

\item Amazon \cite{dou2020enhancing}: It includes product reviews under the Musical Instrument category. The users with more than 80\% of helpful votes were labeled as begin entities, with the users with less than 20\% of helpful votes treated as fraudulent entities.

\item Disney \cite{sanchez2013statistical}: It comes from the Amazon co-purchase network of movies where the attributes are the prices, ratings, number of reviews, etc. The anomalies are labeled manually by majority vote and the ground truths are derived from amazonfail stage information.

\item Amazon-all \cite{dou2020enhancing}:  It includes three types of relations: U-P-U (users reviewing at least one same product), U-S-U (users giving at least one same star rating within one week), and U-V-U (users with top-5\% mutual review similarities). 
Amazon-all is formed by treating the different relations as a single relation following \cite{chen2022gccad, qiao2024truncated}.

\item YelpChi-all \cite{dou2020enhancing}: Similar to Amazon-all, YelpChi-all includes three
types of edges: R-U-R (reviews posted by the same user), R-S-R (reviews for the same
product with the same star rating), and R-T-R (reviews for the same product posted in the
same month). YelpChi-all is formed by treating the different relations as a single relation
following \cite{chen2022gccad, qiao2024truncated}.

\item Tolokers \cite{platonov2023critical}: It is obtained from Toloka crowdsourcing platform where the node represents the person who has participated in the selected project and the edge connects two workers means that they work on the same task. The attributes of the node are the profile and task performance statistics of workers.

\item Question \cite{platonov2023critical}: It is collected from the website Yandex Q where the node represents the user and the edge connecting the node represents the user who has answered other's questions during a one year period. The attribute is the mean of embeddings for words in the description of the user. 
% For the user without description, the additional binary features are employed as the feature of the user.

\item Elliptic \cite{weber2019anti}: It is a Bitcoin transaction network in which the node represents the transactions and the edge is the flow of Bitcoin currency. The Bitcoin transaction is mapped to real-world entities associated with licit categories.

\item T-Finance \cite{tang2022rethinking}: It is a financial transaction network where the node represents an account and the edge represents two accounts that have transaction. The features of each account are related to some attributes of logging like registration days, logging activities, and interaction frequency, etc. The users are labeled as anomalies when they fall into the categories like fraud money laundering and online gambling.

\item  T-Social \cite{tang2022rethinking}: It is a social network where the node represents the user and the edge indicates the relationship between users for more than three months. The attributes of each account are some logging information like registration days, logging activities, interaction frequency, etc.
\end{itemize}

% \section{Description of Baselines}\label{app:competingmethods}
% \subsection{Competing GAD Methods}

% A more detailed introduction of the eight GAD models we compare with is given as follows.

% \subsection{Competing Semi-Supervised Methods}

% \begin{itemize}
% \item DOMINANT \cite{ding2019deep} learns the normal pattern by reconstructing the attribute and structure of the graph using GNN. The anomalies score is the combination of reconstruction errors on the attribute and structure.

% \item  AEGIS \cite{ding2021inductive} synthesizes pseudo-abnormal nodes based on the Gaussian distribution and uses adversarial learning to distinguish between actual abnormal nodes and generated pseudo-abnormal nodes.

% \item  GGAD \cite{qiao2024generative} is the first semi-supervised approach for graph anomaly detection that leverages partially labeled normal nodes. It generates pseudo-anomalous nodes by incorporating two key priors related to graph anomalies to provide effective negative samples for training a discriminative one-class classifier.

% \end{itemize}

\section{Additional Experimental Results}

\subsection{Few-shot Tuning on Abnormal Nodes} \label{app:tuning_abnormal}
Additionally, tuning the abnormal prototype ${\bf{p}_a}$ is also feasible when few-shot abnormal node labels are available. Give some labeled abnormal nodes ${{\mathcal V}_l^{Test}}$, the fine tuning process is formulated as  
\begin{equation}
\label{eq:fine_tune}
{L_{{{pt }}}} = \sum\limits_{i = 1}^{\left| {{\mathcal V}_l^{Test}} \right|} \left\| {{{\bf{r}}_i} - {\bf{p}^\prime_a}} \right\|_2^2,
\end{equation}
where ${{\bf{p}^\prime_a}}  = {{\bf{p}}_a} + {g}({{\bf{p}}_a;\phi}) + \Psi_{\mathbf{p}_a}$,  with a similar formulation applied during the tuning of few-shot normal nodes. To evaluate the performance of AnomalyGFM under few-shot abnormal nodes setting. we include a few labeled abnormal nodes, and treat the unlabeled nodes as normal. These can be combined to serve as the training source. We compare AnomalyGFM with the few-shot GAD method MetaGAD \cite{xu2024metagad} and fully supervised methods BWGNN \cite{tang2022rethinking}. We show results in Table \ref{tab:few_shot_abnormal} for the scenario where few-shot labeled abnormal nodes are available. As shown in Table  \ref{tab:few_shot_abnormal}, AnomalyGFM outperforms MetaGAD and BWGNN since they are limited by the scarcity of anomalies on the target graphs, whereas AnomalyGFM has better generalization since it only needs to adapt the pre-trained anomaly prototype with the few anomaly nodes to the target graphs.

\begin{table}[ht]
\centering
\setlength{\tabcolsep}{1.35mm}
\begin{center}
\caption{Comparison under 1/5-shot anomalies}
\label{tab:few_shot_abnormal}
\scalebox{0.9}{
\begin{tabular}{c|c|c|cc}
\toprule
\hline
\multirow{2}*{\textbf{Metric}} &\multirow{2}*{\textbf{Setting}} & \multirow{2}*{\textbf{Method}}&  \multicolumn{2}{c}{\textbf{Dataset}}\\
&&& Amazon-all &  YelpChi-all\\
\hline
\multirow{6}*{AUROC} & \multirow{3}*{1-shot} 
       &BWGNN (ICML'22) & 0.7441  & 0.5447 \\
     & &MetaGAD (DSAA'24) & 0.8602  & 0.5434 \\
       & &AnomalyGFM &\textbf{0.9155}  & \textbf{0.5941} \\
       \cline{2-5}
      & \multirow{3}*{5-shot} 
       &BWGNN (ICML'22) & 0.8292  & 0.4759 \\
     &  & MetaGAD (DSAA'24)& 0.8558  & 0.5741 \\
       & &AnomalyGFM & \textbf{0.9246}  & \textbf{0.6004} \\
\hline
\multirow{6}*{AUPRC}  &\multirow{3}*{1-shot} 
      & BWGNN (ICML'22) & 0.2947  & 0.1703 \\
      &&MetaGAD (DSAA'24) & \textbf{0.7562}  & 0.1681 \\
       & &AnomalyGFM &0.6956  & \textbf{0.1922} \\
     \cline{2-5}
         & \multirow{3}*{5-shot} 
      &BWGNN (ICML'22) &0.4433  & 0.1408 \\
     &  &MetaGAD (DSAA'24) & \textbf{0.7567}  & 0.1814 \\
       & &AnomalyGFM & 0.7538  & \textbf{0.1935} \\
\hline
\bottomrule
\end{tabular}
}
\end{center}

\end{table}

\subsection{Train the Model on Amazon and Evaluate on Other Datasets}
\label{app:train_amazon}

To further demonstrate the effectiveness of AnomalyGFM, we pre-train the AnomalyGFM on Amazon and evaluate it on the other datasets. The AUROC and AUPRC results are shown in the Table \ref{tab:pre_train}.
AnomalyGFM consistently outperforms competing methods from supervised methods, unsupervised methods, and generalist methods across most datasets in both AUROC and AUPRC, demonstrating its strong generalization when pre-trained on a different auxiliary dataset.

\begin{table*}
\caption{Comparison of AUROC and AUPRC results on 8 real-world GAD datasets under zero-shot setting with the models trained on Amazon only. 
% For each dataset, the best performance per column within each metric is boldfaced, with the second-best underlined. All results are the average of three runs with different random seeds. 
}
% \gs{pls also add avg. and p-value for this table if we want to include it into the appendix.}
\label{tab:pre_train}
\begin{center}
\setlength{\tabcolsep}{0.75mm}
\begin{tabular}{c|c|cccccccccc}
\toprule
\hline
\multirow{2}*{\textbf{Metric}}&\multirow{2}*{\textbf{Method}} & \multicolumn{8}{c}{\textbf{Dataset}} &\multirow{2}*{\textbf{Avg.}} &\multirow{2}*{\textbf{p-value}}\\
&&Facebook &Reddit & Disney &YelpChi-all &Tolokers & Question & Elliptic&T-Finance \\
\hline
\multirow{11}{*}{AUROC}
% \rowcolor{lightgray} 
&\multicolumn{11}{c}{Unsupervised Methods}\\
\cline{2-12}
&AnomalyDAE (ICASSP'20) &0.6123   &\underline{0.5799} & 0.4938  &0.4912 &0.5080  &0.5401   &0.2769   &0.2886  &0.4738 &0.0156 \\
&CoLA (TNNLS'21) & 0.5427   &0.4962  & 0.5455  &0.4937  &0.4680 &0.4768    &\underline{0.6654} &0.4108  &0.5123  &0.0390  \\
& GADAM (ICLR'24) &0.6024  &0.4720  &0.3966 & 0.5289 &\underline{0.5193}  &\textbf{0.5575}   &0.2907 &0.1463  &0.4391 &0.0234    \\
\cline{2-12}
&\multicolumn{11}{c}{Supervised Methods}\\
\cline{2-12}
& BWGNN (ICML'22)  &0.5441 &0.4026  & 0.4196 & 0.5841 &0.4430  &0.5117     &0.5471 &\textbf{0.6196} &0.5089 &0.0390 \\
&XGBGraph (NeurIPS'23)  &0.4869 &  0.4869 & 0.4376   & \underline{0.5869} &0.4710  &\underline{0.5430}    &0.5367 &\underline{0.5950}  & \underline{0.5180} &0.0781 \\
\cline{2-12}
&\multicolumn{11}{c}{Generalist Methods}  \\
\cline{2-12}
&GraphPrompt  (WebConf'23)  & 0.3093  & 0.4511 &\underline{0.7128}  &0.4994 &0.5161  &0.4720    &\textbf{0.6772}  &0.4753 & 0.5141  &0.0546 \\ 
&UNPrompt (Arxiv'24)  & \underline{0.7917} & 0.5356  &  0.6959 &  0.5448&0.4949  &0.4467    &0.3558  &0.1805 &0.5057  &0.0078 \\
\cline{2-12}
&AnomalyGFM   & \textbf{0.8279} & \textbf{0.6331} & \textbf{0.7135}  & \textbf{0.5898} & \textbf{0.5305}  & 0.5268  &0.6010 & 0.5784  &\textbf{0.6251} &/\\

\hline
\multirow{11}{*}{AUPRC}
&\multicolumn{11}{c}{Unsupervised Methods}\\
\cline{2-12}
&AnomalyDAE (ICASSP'20) & 0.0675 & \underline{0.0413}  & 0.0583 &0.1479 &0.2253  &0.0361    &0.0630 &0.0300 &0.0836 &0.0234\\
&CoLA (TNNLS'21)  & 0.0468 & 0.0327  & 0.0717   & 0.1474 &0.2038  &0.0289    &\underline{0.1427}  &0.0367 & 0.0888 &0.0546\\
& GADAM (ICLR’24) & 0.0461 & 0.0299 &0.0732  &0.1602  &\underline{0.2348}  &0.0424    &0.0636 &0.0255  & 0.0844  & 0.0156   \\
\cline{2-12}
&\multicolumn{11}{c}{Supervised Methods}\\
\cline{2-12}
& BWGNN (ICML'22)  & 0.0289& 0.0263  & 0.0494 &  0.1975  &0.2073  &0.0316  &0.0992   &\textbf{0.0635}  &0.0879 & 0.0390\\
&XGBGraph  (NeurIPS'23) &0.0268& 0.0315 & 0.0541  &\textbf{0.1994}  &0.1982  &\textbf{0.0426}    &0.0991 &0.0538  & 0.0881 &  0.0546 \\
\cline{2-12}
&\multicolumn{11}{c}{Generalist Methods} \\
\cline{2-12}
&GraphPrompt  (WebConf'23) & 0.0169 & 0.0298 & \underline{0.1157} & 0.1481  &0.2166   &\underline{0.0425}     &\textbf{0.1590} &0.0464  & 0.0968& 0.1484  \\ 
&UNPrompt (Arxiv'24)  & \textbf{0.2291}& 0.0340   & 0.0933 & 0.1767  &0.2294  &0.0274  &0.0708 &0.0258   &\underline{0.1108} & 0.1953  \\
\cline{2-12}
&AnomalyGFM  & \underline{0.1094}   & \textbf{0.0440 } &\textbf{0.1593}   &\underline{0.1979} & \textbf{0.2505} &0.0322   &0.1112 & \underline{0.0604}  & \textbf{0.1206} & /  \\
\hline 
\bottomrule
\end{tabular}
 
\end{center}
\end{table*}

\subsection{Sensitivity Analysis}  \label{app:sensitive}
% \noindent \textbf{Impact of commonality dimension $d^\prime$.} 
% The AUPRC results are shown in Fig. \ref{fig:dimension_pr}. We can see that AnomalyGFM still outperforms all the competing methods on AUPRC under the different dimensions, indicating that AnomalyGFM is generally robust to the feature dimensionality.

% \begin{figure}
% \setlength{\abovecaptionskip}{0.2cm}
%  \centering
%  % Requires \usepackage{graphicx}
%  \includegraphics[width=3.15in,height=1.1in]{attach/dimension_pr.pdf }\\
%  \caption{AUPRC of AnomalyGFM w.r.t commonality dimension $d^\prime$ 
% }
% \label{fig:dimension_pr}
% \end{figure}

\noindent \textbf{Impact of $\beta$ in Anomaly Scoring.} As shown in Table \ref{tab:scoring},  the set of $\beta$ values are determined based on the prior we gain from the global average similarity in Table \ref{tab:dataset}. It is set to 0 by default for graphs with high global similarity and 4 for graphs with low global similarity. In the few-shot scoring, $\beta$ is set to $0.5$ by default for graphs with high global similarity and four for graphs with low global similarity, allowing the anomaly scores to be computed using both normal and abnormal class prototypes, where the normal class prototype is refined using the labeled normal nodes.

\begin{table}[ht]
\centering
\setlength{\tabcolsep}{2.85mm}
\begin{center}
\caption{The value of $\beta$ in anomaly scoring in both zero-shot and few-shot setting. Sim. indicate the global average edge similarity of the graph.}
\label{tab:scoring}
\begin{tabular}{c|cc}
\hline
Setting& \textbf{Sim.}>0.5 &  \textbf{Sim.}<=0.5\\
\hline
Zero-shot Setting & 0  & 4 \\
\hline
Few-shot Setting  & 0.5 & 4 \\
\hline
\end{tabular}
\end{center}
% \vspace{-1em}
\end{table}

\noindent \textbf{Impact of $\mu$ and $\sigma$ in Gaussian distribution initialization.} To evaluate the effectiveness of AnomalyGFM, we set different values for $\mu$ and $\sigma$ during the Gaussian distribution initialization for both the normal class prototype and the abnormal class prototype, see in the Table \ref{tab:raw_distribution}. The performance remains stable when we adjust $\mu$ and $\sigma$, indicating that AnomalyGFM is not dependent on the initial Gaussian distribution. This is mainly because the supervised loss function provides sufficient supervision, reducing the reliance on the Gaussian distribution.

\begin{table}[ht]
\caption{Performance of AnomalyGFM under different initialization hyperparameters for Gaussian distribution}
\setlength{\tabcolsep}{0.85mm}
\label{tab:raw_distribution}
\begin{center}
\scalebox{0.85}{
\begin{tabular}{c|c|ccccc}
\toprule
\hline
\multirow{2}*{\textbf{Metric}}&\multirow{2}*{\textbf{Initialization}} & \multicolumn{5}{c}{\textbf{Dataset}}\\
&      & Amazon-all   & Tolokers & Question & Elliptic &  T-Finance   \\
% \midrule
\hline
\multirow{4}*{AUROC}
& $\mu$ = 0, $\sigma$ = 1     &  0.9032 &  0.5843&  0.5280  & 0.6195  &  0.5614    \\
& $\mu$ = 0, $\sigma$ = 0.5      &   0.9229  & 0.5981   & 0.5378 &  0.6068 & 0.5902  \\
& $\mu$ = 0.1, $\sigma$ = 0.5  &   0.8756   & 0.6004  &  0.5437 & 0.5907  &  0.5727 \\
& $\mu$ = 0.2, $\sigma$ = 1     & 0.8963   & 0.5975   & 0.5429  &   0.6099& 0.5922  \\
\cline{2-6}
\hline
\hline
\multirow{4}*{AUPRC}
& $\mu$ = 0, $\sigma$ = 1    &  0.6820   & 0.2749  &  0.0335 &  0.1333 & 0.0593  \\
& $\mu$ = 0, $\sigma$ = 0.5     & 0.7334    &  0.2824 & 0.0345 &   0.1234  &0.0612  \\
& $\mu$ = 0.1, $\sigma$ = 0.5    &  0.6568   &  0.2870 &  0.0342&  0.1096 &  0.0596  \\
& $\mu$ = 0.2, $\sigma$ = 1 &   0.6959   &  0.2853  & 0.0339  &0.1177   &  0.0604  \\
\cline{2-6}
\hline
\bottomrule
\end{tabular}}
\end{center}
% \vspace{-1em}
\end{table}

\section{Algorithms}\label{algo}
The pre-training, zero-shot inference, few-shot prompt fine-tuning, and subgraph-based inference processes of AnomalyGFM are summarized in Algorithm \ref{alg:pretrain}, Algorithm \ref{alg:zero_shot}, Algorithm \ref{alg:few_shot}, and Algorithm \ref{alg:subgraph}, respectively.

\begin{algorithm}[H]
\small
\caption{Pre-training of AnomalyGFM} 
\begin{algorithmic}[1]
\label{alg:pretrain}
\STATE {\textbf{Input:}} Training graph  $\mathcal{G}_{\rm{Train}}= ({\mathcal V_{train}, \mathcal E_{train}})$; Pre-training epoch $E$.\\
\STATE {\textbf{Output:}} Pre-training $AnomalyGFM(\mathbf{A}, \mathbf{X}).$
\STATE Perform feature unification of $\mathbf{X}$ to obtain $\tilde{\mathbf{X}}$.
\STATE {Initialize the Gaussian distribution  $\bf{z}_n$ and $\bf{z}_a$  for normal class prototype and abnormal class prototype.}
\STATE {Randomly initialize GNN ${{(\mathbf{h}^{(0)}_1, \mathbf{h}^{(0)}_2,...,\mathbf{h}^{(0)}_N)} \leftarrow {{\tilde{\bf{X}}}}}$ }
\FOR{ ${epoch = 1, \cdots ,E}$ }
{
\FOR{each ${v}$ in ${{\mathcal V_{train}}}$}
{
 \FOR{ ${l = 1, \cdots ,L}$}{
   \STATE{${{\mathbf{h}}_{v}^{(l)}  =  {{\mathbf{W}}^{(l)}\mathbf{h}}_{v}^{(l-1)} }$}

 \STATE{${{\mathbf{h}}_{v}^{(l)} = {\mathop{\rm R}\nolimits} {\rm{eLU}}\left( {{\rm{AGG}}(\{ {\mathbf{h}}_{v'}^{(l)} :(v,v') \in {\mathcal E}_{train}\} )} \right)}$ }

 }\ENDFOR
   }\ENDFOR
    \STATE{Compute the residual feature ${\bf{r}}_i$ of each node $v_i$ using Eq. (\ref{eq:residual}).}
    \STATE{Compute the ${L_{BCE}}$ for classifier ${{p_i} = {f_\theta }\left( {\bf h}_i \right)}$.}
  \STATE{Alignment the normal class prototype and abnormal class prototype with the residual feature using  ${L_{Alignment}}$}
    \STATE{Compute the total loss ${L_{{total}}} = {L _{BCE}} + \alpha {L _{Alignment}}$.} 
\STATE{Update the trainable weight parameters $\bf{W}$, ${\theta}$, $\Theta_a$ and $\Theta_n$ by using gradient descent.} 

}\ENDFOR

 \RETURN Pre-trained $AnomalyGFM(\mathbf{A}, \mathbf{X})$, $\Phi({\bf{z}}_n; \Theta_n)$ and  $\Phi({\bf{z}}_a; \Theta_a).$

\end{algorithmic}
\end{algorithm}

\begin{algorithm}[H]
\small
\caption{Zero-shot Inference of AnomalyGFM}
\begin{algorithmic}[1]
\label{alg:zero_shot}
\STATE {\textbf{Input:}} Testing graph $\mathcal{G}_{\text{test}} = ({\mathcal V_{test}, \mathcal E_{test}})$, pre-trained  $AnomalyGFM(\mathbf{A}, \mathbf{X})$,  $\Phi({\bf{z}}_n; \Theta_n)$ and  $\Phi({\bf{z}}_a; \Theta_a)$.
\STATE {\textbf{Output:}} Anomaly score of testing nodes.
% \STATE {Initialization of the Gaussian distribution of normal class prototype and abnormal class prototype.}
 \FOR  {each $v_i$ in ${{\mathcal V_{test}}}$}
  {
   \STATE {Apply the pre-trained $AnomalyGFM(\mathbf{A}, \mathbf{X})$ on the graph.}
   \STATE{Compute the residual feature ${\bf{r}}_i$ of each node $v_i$ using Eq. (\ref{eq:residual}).}
    \STATE{Scoring using Similarity measurement with normal class prototype ${\bf p}_n$ and abnormal class prototype ${\bf p}_a$} 
    \STATE{Compute the anomaly score using Eq. (\ref{eq:scoring}).}
}
\ENDFOR
 \RETURN Anomaly scores of testing nodes ${s(v_1), \cdots ,s(v_N)}$.
\end{algorithmic}
\end{algorithm}

\begin{algorithm}[H]
\small
\caption{Few-shot Prompt Fine-tuning of AnomalyGFM}
\begin{algorithmic}[1]
\label{alg:few_shot}
\STATE {\textbf{Input:}} Testing graphs  $\mathcal{G}_{\text{test}} = ({\mathcal V_{test}, \mathcal E_{test}})$, pre-trained  $AnomalyGFM(\mathbf{A}, \mathbf{X})$,  $\Phi({\bf{z}}_n; \Theta_n)$ and  $\Phi({\bf{z}}_a; \Theta_a)$, the training epoch $e$ of fine-tuning.
\STATE {\textbf{Output:}} The fine-tuned  $AnomalyGFM(\mathbf{A}, \mathbf{X})$ with updated normal class prototype ${\bf p}_n^\prime$.
% \STATE {Initialization of the Gaussian distribution of normal class prototype and abnormal class prototype.}
% \STATE {Randomly initialize GNN ${{(\mathbf{h}^{(0)}_1, \mathbf{h}^{(0)}_2,...,\mathbf{h}^{(0)}_N)} \leftarrow {{\tilde{\bf{X}}_{test}}}}$ }
\FOR{ ${epoch = 1, \cdots e}$ }
{
\FOR{each ${v}$ in ${{\mathcal V_{test}^l}}$}
{
 \FOR{ ${l = 1, \cdots ,L}$}{
   \STATE{${{\mathbf{h}}_{v}^{(l)}  =  {{\mathbf{W}}^{(l)}\mathbf{h}}_{v}^{(l-1)} }$}

 \STATE{${{\mathbf{h}}_{v}^{(l)} = {\mathop{\rm R}\nolimits} {\rm{eLU}}\left( {{\rm{AGG}}(\{ {\mathbf{h}}_{v'}^{(l)} :(v,v') \in {\mathcal E}_{test}\} )} \right)}$ }

 }\ENDFOR

 \STATE{Compute the residual feature ${\bf r}_i$ of each labeled normal node $v_i$ using Eq. (\ref{eq:residual})}
    }\ENDFOR
 \STATE{Compute the $L_{pt}$ and optimize the newly added trainable parameters $\phi, \Psi$ using gradient descent.}

}\ENDFOR

\RETURN The fine-tuned $AnomalyGFM(\mathbf{A}, \mathbf{X})$ with the updated normal class prototype ${\bf p}_n^\prime$.

\end{algorithmic}
\end{algorithm}

\begin{algorithm}[ht]
\small
\caption{Subgraph-based Inference of AnomalyGFM}
\begin{algorithmic}[1]
\label{alg:subgraph}

\STATE {\textbf{Input:}} Testing graphs  $\mathcal{G}_{\text{test}} = ({\mathcal V_{test}, \mathcal E_{test}})$,  pre-trained  $AnomalyGFM(\mathbf{A}, \mathbf{X})$,  $\Phi({\bf{z}}_n; \Theta_n)$ and  $\Phi({\bf{z}}_a; \Theta_a)$.
\STATE {\textbf{Output:}} Anomaly score of testing nodes.
 \FOR  {each $v_i$ in ${{\mathcal V_{test}}}$}
  {
  \STATE {Extract the subgraph ${\mathcal{S}}\left(v_i\right)$ of node $v_i$ using random walk.}
   \STATE {Apply the pre-trained $AnomalyGFM(\mathbf{A}, \mathbf{X})$ on the extracted subgraph.}
   \STATE{Compute the residual feature ${\bf{r}}_i$ of each node $v_i$ using Eq. \ref{eq:residual_subgraph}.}
    \STATE{Similarity measurement with normal class prototype ${\bf p}_n$ and abnormal class prototype ${\bf p}_a$.} 
    \STATE{Compute the anomaly score using Eq. (\ref{eq:scoring}).}
}
\ENDFOR
 \RETURN Anomaly scores of testing nodes ${s(v_1), \cdots ,s(v_N)}$.
\end{algorithmic}
\end{algorithm}

\end{document}